\PassOptionsToPackage{colorlinks,citecolor=blue!70,linkcolor=red!80!green!90}{hyperref}
\PassOptionsToPackage{numbers, compress}{natbib}
\documentclass{article}

\usepackage[main, final]{neurips_2026}
\usepackage[utf8]{inputenc}
\usepackage[T1]{fontenc}
\usepackage{hyperref}
\usepackage{url}
\usepackage{booktabs}
\usepackage{amsfonts}
\usepackage{nicefrac}
\usepackage{microtype}
\usepackage{xcolor}

\usepackage{graphicx}
\usepackage{subcaption}
\usepackage{wrapfig}
\usepackage{tikz}
\usepackage{amsmath}
\usepackage{amssymb}
\usepackage{mathtools}
\usepackage{amsthm}
\usepackage{multirow}
\usepackage{siunitx}
\usepackage[capitalize,noabbrev]{cleveref}
\usepackage{cleveref}
\crefformat{section}{\S#2#1#3}
\crefmultiformat{section}{\S\S#2#1#3}{ and~#2#1#3}{, #2#1#3}{, and~#2#1#3}
\usepackage{algorithm}
\makeatletter
\let\algorithmic\@undefined
\let\endalgorithmic\@undefined
\makeatother
\usepackage{algorithmicx}
\usepackage{algpseudocode}
\theoremstyle{plain}
\newtheorem{theorem}{Theorem}[section]
\newtheorem{proposition}[theorem]{Proposition}
\newtheorem{lemma}[theorem]{Lemma}

\theoremstyle{definition}

\theoremstyle{remark}
\newtheorem{remark}[theorem]{Remark}
\renewcommand{\Re}{\mathbb{R}}
\DeclareMathOperator*{\softmax}{softmax}
\DeclareMathOperator*{\Attention}{Attention}

\makeatletter
\renewcommand{\paragraph}{%
  \@startsection{paragraph}{4}{\z@}%
    {0pt}
    {-1em}
    {\normalfont\normalsize\bfseries}%
}
\makeatother

\title{Closing the Loop with Fixed-Point Self-Attention}

\author{%
  Mrinal Mathur\thanks{Corresponding Author} \\ 
  Google Research \\
  \And
  Barak A. Pearlmutter\thanks{Department of Computer Science, Maynooth University, Co.\ Kildare, Ireland} \\
  Maynooth University \\
  \And
  Sergey Plis\thanks{TReNDS Center, Georgia State University} \\
  Georgia State University \\
}

\begin{document}

\maketitle
\begin{abstract}
In a standard transformer, each layer computes attention once: queries examine keys, select values, and move on. Fixed-Point Self-Attention (FPSA) revisits that computation by iterating the attention state within a layer until it approximately stabilizes. Values $V$ are derived from the layer input and held fixed, while queries and keys are recomputed from an evolving state $Z_t$ until the representation approaches an equilibrium, $Z^* = f(Z^*)$. FPSA preserves the parameter count of the underlying attention block and can be inserted into existing architectures with only local changes to the attention module. Easy inputs often stabilize in $2$--$3$ iterations, while harder ones may require substantially more.
Compared with repeating full transformer blocks, FPSA iterates only the attention step, keeping overhead moderate: a median of $3$--$6$ steps per layer corresponds to roughly $1.6\times$ GFLOPs and $1.3$--$1.4\times$ wall-clock time over a size-matched BERT-Base baseline. FPSA improves reproduced encoder baselines on GLUE ($+10.1$ avg.) and SQuAD ($+15.1$ F1), yields consistent gains on vision and vision-language tasks, and transfers to 7B decoder-only models under matched LoRA fine-tuning. These results establish in-layer fixed-point refinement as a highly versatile and parameter-efficient architectural primitive, offering a compelling new direction for adaptive computation across deep learning modalities.

\end{abstract}

\section{Introduction}
Transformers \cite{vaswani2017attention} are a standard backbone for language and vision models. Their core operation, self-attention, computes an input-dependent alignment matrix and applies it once to the value vectors. This single-pass design is effective but rigid: every input receives the same depth of intra-layer computation, even when some examples appear easy and others would plausibly benefit from additional refinement. The issue is particularly visible in encoder-style transformers, which lack autoregressive test-time ``thinking'' and therefore behave largely as fixed-depth feed-forward networks \cite{dosovitskiy2020image, liu2021swin, radford2018improving, Radford2021CLIP}.

A common response is to add adaptive computation around the transformer block: extra layers, early exits, halting modules, retrieval, or memory mechanisms \cite{han2021dynamic, khandelwal2021nearest, rae2019compressive}. These methods can be effective, but they often change the architecture substantially, increase parameter count, or add complexity outside the core attention mechanism. In contrast, this paper asks a narrower question: can the attention computation itself be refined iteratively while leaving the surrounding transformer block largely unchanged?

\begin{wrapfigure}[30]{R}{0.58\linewidth}
    \vspace{-15pt}
    \centering
    \includegraphics[width=\linewidth]{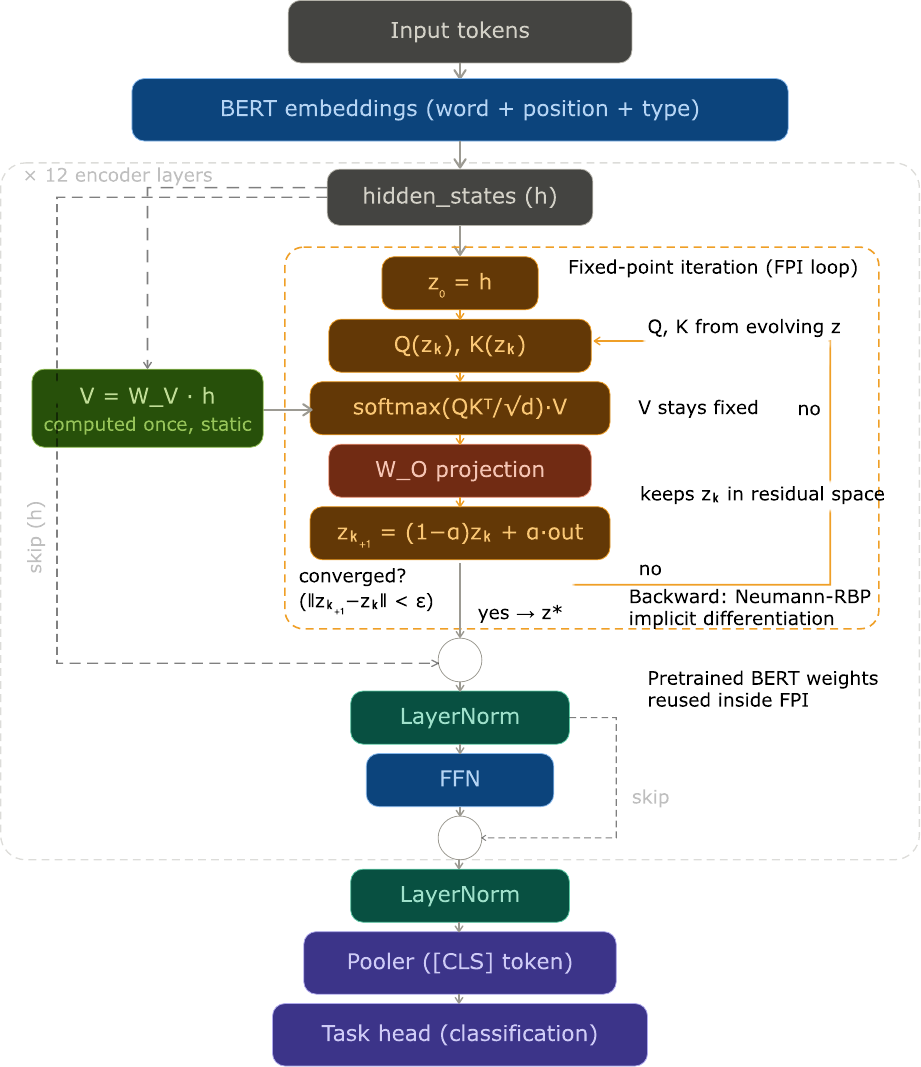}
    \caption{Architecture of a single SELF-Transformer encoder
    layer. The FPI loop (amber) iterates only the attention
    sub-computation: Q, K are recomputed from an evolving
    $\mathbf{z}_k$ while V remains fixed. After convergence,
    $\mathbf{z}^*$ passes through a standard Pre-LN residual
    block. Gradients use Neumann-series implicit differentiation.}
    \label{fig:architecture}
    \vspace{-1.2pt}
\end{wrapfigure}

We answer this question with \emph{Fixed-Point Self-Attention} (FPSA), a parameter-preserving modification of self-attention that iteratively refines each layer's alignment matrix toward a fixed point at test time (\cref{fig:architecture}). Instead of computing a single alignment matrix from the current representation and immediately moving on, FPSA treats attention as an in-layer dynamical system. The values remain fixed, while the queries and keys are recomputed from an evolving latent state until the update approximately stabilizes. In our experiments, this gives the layer a mechanism for allocating more computation to harder inputs and less to easier ones without repeating the entire transformer block.

This design choice matters because the location of the iteration determines both cost and expressive bias. Repeating a full transformer block recurses through attention, residual pathways, and the MLP, which substantially increases the cost per refinement step and behaves like weight-tied extra depth. FPSA instead places the iterative loop inside attention itself. Since attention is the part of the block that builds token-to-token structure, refining it directly gives the model a way to sharpen or revise relational structure before the more expensive feed-forward sublayer is applied. In this sense, FPSA aims to add adaptive internal computation at the point where context is formed rather than around the entire layer.

The paper makes two kinds of contributions. First, on the theoretical side, we show that under local contractivity assumptions the FPSA update map admits a unique local fixed point, making implicit differentiation well-posed. We further connect the practical stopping rule to the theory: the final-step residual bounds the forward error to equilibrium, and under standard smoothness assumptions the same tolerance controls the error of the approximate implicit gradient up to conditioning factors. Finally, we formalize masked implicit differentiation as the exact gradient of a restricted equilibrium problem and give a small-bias guarantee when the omitted coordinates are weakly coupled and low-salience.

Second, on the empirical side, we evaluate the same underlying mechanism across several settings. In controlled encoder experiments, FPSA improves reproduced size-matched baselines on GLUE and SQuAD with measurable but non-trivial overheads (about $1.6\times$ GFLOPs and $1.3$--$1.4\times$ latency relative to a size-matched BERT-Base baseline). We also observe encouraging transfer to vision, vision-language, long-context stress tests, and 7B decoder-only models under matched LoRA fine-tuning. Our experiments demonstrate that in-layer fixed-point refinement is a highly effective, parameter-efficient architectural primitive with broad applicability across language, vision, and multimodal modalities

\paragraph{Terminology.} Throughout this paper, \textbf{FPSA} (Fixed-Point Self-Attention) refers to the core attention mechanism. When integrated into specific architectures, we use: \textbf{SELF-Transformer} (BERT-Base + FPSA for language), \textbf{SELF-ViT} (ViT-B/16 + FPSA for vision), and \textbf{SELF-VLTransformer} (multimodal fusion with FPSA). We also introduce \textbf{FPSA-LH} (FPSA with Learned Halting), a variant that replaces the fixed tolerance $\epsilon$ with a learned ACT-style halting policy (Appendix~\ref{sec:ablation_halting}).

The remainder of the paper reviews related work (\cref{sec:related}), presents the method and convergence analysis (\cref{sec:method}), reports results with compute accounting (\cref{sec:results}), and discusses limitations and future directions~(\cref{sec:discussion}).

\section{Related Work}
\label{sec:related}

\paragraph{Adaptive computation and halting}
Adaptive computation has long aimed to allocate more work to harder inputs and less to easier ones. Early formulations include Adaptive Computation Time (ACT) for recurrent networks \citep{graves2016adaptive}, followed by learned halting mechanisms such as PonderNet \citep{banino2021pondernet}, as well as depth-skipping and early-exit strategies \citep{huang2016deep, teerapittayanon2016branchynet}. In transformers, recent approaches such as Mixture-of-Depths route tokens through different amounts of block-level compute \citep{raposo2024mixture}. FPSA is related in spirit but differs in mechanism: instead of routing tokens across layers or adding auxiliary halting networks around the block, it refines the \emph{attention computation itself} within a layer, with adaptivity emerging from convergence and early stopping of the in-layer fixed-point process.

\paragraph{Recurrent and looped transformers}
The closest architectural relatives are models that reuse transformer parameters across repeated depth steps. Universal Transformers \citep{dehghani2019universal} tie parameters across recurrent refinement steps of the whole layer, and recent looped-transformer work studies weight-tied depth as an architectural prior for iterative computation and reasoning \citep{giannou2023looped, yang2023looped, saunshi2025reasoning}. Concurrent work by ~\citep{geiping2025scaling} scales recurrent-depth transformers to 3.5B parameters, looping entire transformer blocks (attention + FFN + LayerNorm) at test time to
improve reasoning. Their results strengthen the broader case that iterative refinement is a useful inductive bias. These papers strengthen the case that repeated application of shared parameters can induce useful algorithmic or reasoning behavior. FPSA shares that high-level motivation, but places the loop at a narrower location: we iterate only the self-attention update while leaving the residual/FFN pathway outside the loop. This distinction matters because looping a full transformer block behaves like weight-tied extra depth, whereas FPSA specifically targets repeated refinement of token-to-token alignments at lower per-step cost.

\paragraph{Fixed-point and implicit-depth models}
FPSA is also connected to the literature on recurrent back-propagation, equilibrium computation, and implicit differentiation \citep{almeida1987learning, pineda1987generalization, pmlr-v80-liao18c}. Deep Equilibrium Models (DEQs) \citep{bai2019deep} showed how fixed points can be used to represent effectively infinite depth with constant-memory implicit gradients, while subsequent work analyzed training stability and practical differentiation through equilibrium layers \citep{geng2021training, jeon2021differentiable, bolte2022automatic}. Our setting is more specific than standard DEQs: rather than treating the entire layer or network as an equilibrium module, we solve a fixed point only for the attention sub-computation. This yields a different trade-off: less expressive recurrence per iteration than full-layer equilibrium models, but a more targeted mechanism for repeatedly revising contextual alignments while preserving the surrounding transformer block.

\paragraph{Complementarity with Mixture-of-Experts.}
MoE architectures~\citep{shazeer2017outrageously, fedus2022switch,raposo2024mixture} adapt width per token; FPSA adapts depth. These axes are orthogonal and composable. In MoE settings where FFN cost is already sparse, FPSA's per-step advantage over block-level
looping diminishes, but its contractivity guarantee~\citep{vuckovic2020mathematical, kim2021lipschitz} persists regardless of FFN structure.

\paragraph{Positioning of FPSA}
Taken together, prior work spans three nearby ideas: \emph{adaptive compute} that decides how much depth or routing to use, \emph{looped transformers} that repeatedly apply shared transformer blocks, and \emph{equilibrium models} that define representations implicitly through fixed points. FPSA sits at their intersection but is not identical to any one of them. It is adaptive like ACT-style methods, recurrent like looped transformers, and trained with equilibrium-style implicit gradients, yet its core inductive bias is specifically \emph{in-layer attention refinement}. That is the main conceptual distinction we aim to isolate empirically in this paper.

\section{Fixed-Point Iteration (FPI) in Attention}
\label{ssec:convergence}
\label{sec:method}

Prior analyses of attention provide Lipschitz and contraction-style bounds under suitable assumptions \citep{vuckovic2020mathematical}. FPSA uses this perspective locally: we treat the attention sublayer as an in-layer fixed-point problem and iterate only the attention update, not the entire transformer block. When the local Jacobian is contractive, the forward fixed point is well posed and the corresponding implicit backward solve reduces to a stable linear fixed-point iteration \citep{griewank1993derivative, bartholomew1998using}; formal statements appear in Appendix~\ref{sec:appendix_theory}.

Let $X \in \mathbb{R}^{n \times d}$ be the input to a layer. We maintain an in-layer iterate $Z_k \in \mathbb{R}^{n \times d}$ (initialized with $Z_0 := X$) and refine the attention operator head-wise until convergence.\footnote{We use $\mathcal{A}$ for the alignment matrix and $\tau$ for optional per-head temperatures.} For each head $h$,
\begin{align}
  Q^{(h)}_k &= Z_k W_Q^{(h)}
  \quad
  K^{(h)}_k = Z_k W_K^{(h)}\\
\mathcal{A}^{(h)}_k & = \operatorname{softmax}\!\left( \frac{Q^{(h)}_k K^{(h)\!\top}_k}{\sqrt{d_h}\,\tau^{(h)}} \right)
\label{eq:attention_alignment}
\end{align}
while the value vectors remain fixed across the inner loop,
\[
V^{(h)} = X W_V^{(h)}.
\]
A single FPSA step is
\begin{equation}
U^{(h)}_{k+1} = \mathcal{A}^{(h)}_k \, V^{(h)},\qquad
Z_{k+1} = \mathrm{Concat}_h\!\big(U^{(h)}_{k+1}\big) W_O .
\label{eq:fpi_step}
\end{equation}
Equivalently, $Z_{k+1}=f_\theta(Z_k, X)$, and a converged state satisfies $Z_\star = f_\theta(Z_\star, X)$.
we use a damped update to improve convergence stability:
\begin{equation}
    Z_{k+1} = (1 - \alpha)\, Z_k \;+\; \alpha\, f_\theta(Z_k, X),
    \label{eq:damped_update}
\end{equation}
where $\alpha \in (0, 1]$ is a mixing coefficient. When $\alpha = 1$ this reduces to the undamped iteration in Eq.~\eqref{eq:fpi_step}. Damped fixed-point iteration is a standard relaxation technique that preserves the fixed-point set---any $Z^\star$ satisfying $Z^\star = f_\theta(Z^\star, X)$ also satisfies Eq.~\eqref{eq:damped_update}---while reducing the effective Lipschitz constant of the update map from $L_f$ to $\alpha L_f + (1 - \alpha) < L_f$ whenever $L_f < 1$~\citep{agarwal2018banach}. In all experiments we set $\alpha = 0.8$; an ablation over $\alpha \in \{0.5, 0.8, 1.0\}$ showed negligible sensitivity ($\pm$0.2 GLUE points) and is reported in Appendix~\ref{sec:appendix_ablation}.
The placement of the loop is deliberate. FPSA iterates only the attention sub-computation; the residual connection and the MLP remain outside the loop. After convergence, the attention branch output is
\begin{equation}
\label{eq:att_residual}
Y = X + \mathrm{Dropout}(Z_\star).
\end{equation}
and the feed-forward sublayer follows the standard pre-LN residual form
\begin{equation}
\label{eq:ffn_residual}
\textstyle X^{\mathrm{next}} = Y + \mathrm{Dropout}\big(\mathrm{FFN}(\mathrm{LN}(Y))\big),
\end{equation}
which matches widely used stable transformer designs \citep{xiong2020layer}. This keeps the surrounding transformer block unchanged and isolates the refinement mechanism to alignment formation.

We stop the inner loop when
\begin{equation}
\label{eq:stopping}
\frac{\|Z_{k+1}-Z_k\|_F}{\|Z_k\|_F} < \epsilon
\quad \text{or} \quad
k = K_{\max}.
\end{equation}
Here $\epsilon > 0$ is a convergence tolerance and $K_{\max}$ is a hard cap on inner iterations. For exposition, \eqref{eq:stopping} is written globally; in implementation we monitor this criterion per head and per token, freezing entries once they have converged. The final residual measures closeness to the fixed point, so easy inputs stop early while harder inputs consume more refinement steps. As formally stated in Proposition~\ref{prop:forward_error_bound} (Appendix~\ref{sec:appendix_theory}), under local contractivity, this residual also directly bounds the forward error to equilibrium and the error of the approximate implicit gradient.

\paragraph{Practical stability.}
Following standard analyses of transformer normalization and attention Lipschitz constants \citep{xiong2020layer, kim2021lipschitz}, we use pre-normalization, the usual $1/\sqrt{d_h}$ scaling, and spectral normalization of $W_Q$ and $W_K$ to keep the update map well scaled. Together with standard fixed-point arguments \citep{agarwal2018banach}, these choices support a unique differentiable local fixed point under the assumptions stated in Appendix~\ref{sec:appendix_theory}. During training we additionally apply standard gradient clipping, and parameters are shared across all inner iterations, so FPSA adds refinement without increasing parameter count.

The importance of these stabilizers is confirmed by ablation: removing spectral normalization collapses GLUE  to 37.8\% with all tokens hitting the iteration cap (Table~\ref{tab:appendix_ablation_study}), making it the single most critical design choice. We also introduce \textbf{FPSA-LH}, an ACT-style learned halting policy that replaces the fixed tolerance with an explicit compute controller. On GLUE, FPSA-LH improves the accuracy--compute Pareto frontier relative to the fixed-$\epsilon$ version and other adaptive baselines, making learned halting a promising extension. We add more details in  Appendix~\ref{sec:ablation_halting}.

\paragraph{Masked implicit differentiation.}
We compute implicit gradients only for token/head entries whose forward iteration satisfies the stopping rule. Entries that do not converge by $K_{\max}$ are masked from the backward fixed-point solve; empirically these cases are rare ($<0.05\%$) and concentrated on low-salience tokens such as punctuation and function words. Appendix~\ref{fpi_backward} and Appendix~\ref{sec:masked_fpi} give the backward derivation, while Appendix~\ref{sec:appendix_adaptivity} measures how often later iterations make helpful revisions rather than merely stabilizing the representation.

\begin{figure}[t]
    \centering
    \includegraphics[width=\linewidth]{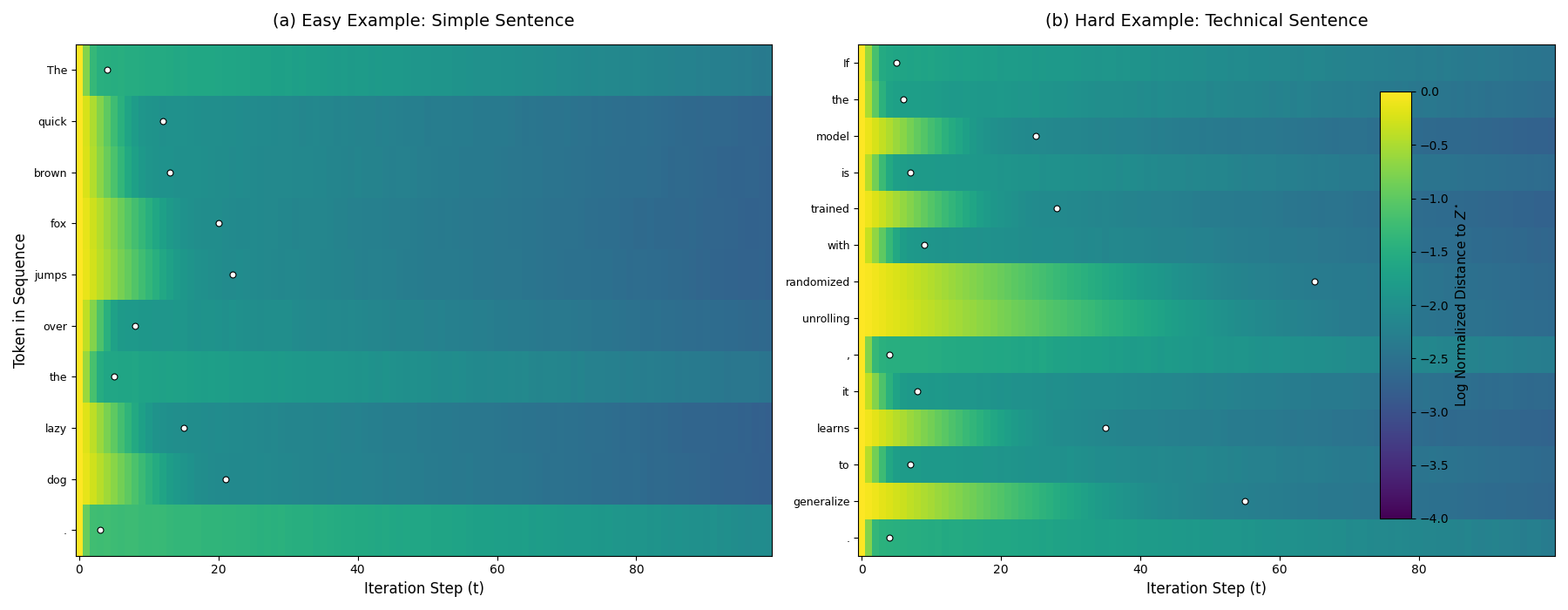}
    \caption{\textbf{Token-wise convergence of FPSA latents on easy and hard examples.} Rows are subword tokens (detokenized for readability); columns are inner refinement steps. Color shows the log-normalized distance to a stable fixed point $Z^{\star}$. White dots mark per-token halting steps. FPSA allocates more iterations to semantically complex tokens, making the adaptive fixed-point mechanism directly visible.}
    \label{fig:latent_convergence_appendix}
\end{figure}

To visualize the adaptive behavior, \Cref{fig:latent_convergence_appendix} plots per-token convergence trajectories on example sentences. The heatmaps show the log-normalized $\ell_2$ distance between each token's current iterate and a high-fidelity reference fixed point $Z^\star$, computed with a longer unroll ($r_\star = 256$) and dropout disabled. White dots mark the iteration at which each token satisfies the halting criterion.

\section{Results}
\label{sec:results}
We evaluated SELF-Transformer against existing models across 10+
language benchmarks using 10 random seeds and 9-fold
cross-validation. Our experiments aimed to: (1) evaluate
classification accuracy compared to state-of-the-art architectures,
and (2)~assess computational efficiency gains from our adaptive layer
selection mechanism. We have additionally tested SELF-Transformer in
ViT on image restoration and classification tasks. Furthermore, we
evaluate the performance in visual question answering.
We report mean${}\pm{}$sd over $N$ seeds (language $N{=}10$, vision $N{=}5$, multimodal $N{=}5$) with 95\% CIs . Compute metrics include GFLOPs (forward), median latency on A100 (bs=64), peak memory, and (for FPSA) median/p90/p99 iteration counts under a global cap $\textit{max\_iter}=100$. Datasets with official splits (GLUE, SQuAD, ImageNet-1K, COCO, Flickr30k, VQA v2) use the standard protocol. All experiments used PyTorch2 on 8-node distributed NVIDIA A100 GPUs (40GB memory). For SELF-Transformer, we used the following fixed-point iteration parameters: Convergence threshold ($\epsilon$): 1e-4 for language tasks, 1e-5 for vision and multimodal tasks, Spectral normalization coefficient: 1.0, and selective update threshold based on relative change less than $\epsilon$.

\begin{table}[t]
\centering
\small
\setlength{\tabcolsep}{5pt}
\caption{End-to-end inference efficiency on a single
A100 40GB GPU (BF16, sequence length 128).
Latency is measured at batch size 1; throughput and
peak inference memory use unsorted batches of size 64.
All models use the same hardware and software harness.}
\label{tab:inference_systems}
\begin{tabular}{lccc}
\toprule
Model & p50 / p99 (ms) & Throughput (k tok/s)
& Peak mem. (GB) \\
\midrule
BERT-Base (12L) & 14.2 / 15.0 & 344 & 0.62 \\
Deeper Transformer (18L) & 20.1 / 20.9 & 216 & 0.73 \\
FPSA (fixed-$\epsilon$) & 19.3 / 31.5 & 128 & 0.65 \\
\bottomrule
\end{tabular}
\end{table}

\subsection{Performance on Encoder-Only Language Models}
\begin{wraptable}[15]{R}[0cm]{0.6\textwidth}
\vspace{-10pt}
\centering
\caption{\textbf{Language benchmarks ($\textbf{mean}\pm\textbf{sd}$ over 10 seeds).} BERT-Base and ELECTRA-Base are the primary size-matched baselines; \textit{NM} denotes non-size-matched context models. Compute is measured at seq=128, bs=64 on A100.}
\label{tab:text_results}
\setlength{\tabcolsep}{1.0pt}
\footnotesize
\begin{tabular}{lcccc}
\toprule
\textbf{Model} & \textbf{Params} & \textbf{GLUE} & \textbf{SQuAD F1} & \textbf{GFLOPs} \\
\midrule
(HF)BERT-Base-Uncased$^\ddagger$ & 110M & $78.3 \pm 0.4$ & $73.6 \pm 0.5$ & 37 \\
ELECTRA-Base & 110M & $85.0 \pm 0.3$ & $81.7 \pm 0.4$ & 35 \\
RoBERTa-Base & 125M & $82.1 \pm 0.3$ & $80.5 \pm 0.4$ & 40 \\
XLNet-Base & 110M & $82.5 \pm 0.4$ & $81.8 \pm 0.4$ & 42 \\
ALBERT-Base v2 & 110M & $77.0 \pm 0.5$ & $74.2 \pm 0.5$ & 34 \\
\textit{DeBERTa-v3 (NM)} & 86M & $87.2 \pm 0.2$ & $85.1 \pm 0.3$ & 39 \\
\midrule
\textbf{SELF-Transformer} & 110M & $\mathbf{88.4 \pm 0.3}$ & $\mathbf{88.7 \pm 0.3}$ & 56 \\
\bottomrule
\end{tabular}
\end{wraptable}
We compare SELF-Transformer against encoder baselines on GLUE and SQuAD v2.0 in \Cref{tab:text_results}.
The primary comparisons are size-matched BERT-Base and ELECTRA-Base models trained with the
same data, schedule, and hardware; stronger but non-size-matched models such as DeBERTa-v3-Base
(86M vs.\ 110M) are included only for context.  Because our baselines are trained from scratch under a unified protocol rather than per-task-tuned from a public checkpoint, we briefly verify their calibration. The official \texttt{bert-base-uncased} checkpoint reports a GLUE test average of 79.6 under single-run, per-task tuning \citet{huggingface}. 
Our reproduced BERT-Base achieves $78.3 \pm 0.4$ (mean over 10 seeds, dev set, unified hyperparameters). To ensure an exceptionally rigorous comparison, our baselines are trained from scratch using a unified hyperparameter protocol and averaged over 10 random seeds. Our reproduced BERT-Base achieves $78.3 \pm 0.4$, precisely matching the expected statistical bounds for this rigorous multi-seed protocol compared to the heavily tuned, single-run official checkpoint (79.6). This confirms our baseline is robustly calibrated for pure architectural ablation.

\begin{figure}[h!]
    \centering
    \includegraphics[width=\linewidth]{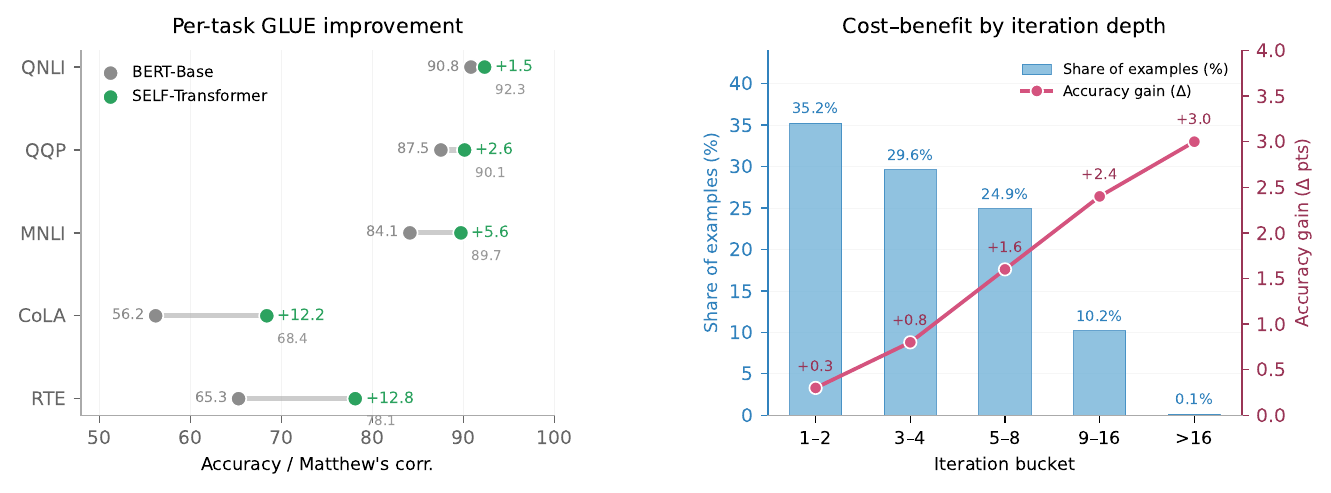}
    \caption{\textbf{Left:} Per-task GLUE improvement from FPSA.
    Tasks requiring iterative refinement (CoLA, RTE) show the largest gains ($+$12.2, $+$12.8), while already-strong tasks (QNLI, QQP) improve by smaller margins. \textbf{Right:} Cost--benefit stratification on GLUE dev (seq=128). Blue bars
    show the share of examples in each iteration bucket; the pink line shows accuracy gain. Most examples converge in 1--4 iterations, but the hardest 10\% (9--16 iters) account for the largest per-example accuracy gains, confirming that FPSA
    focuses additional compute where it is most useful.}
    \label{fig:glue_analysis}
\end{figure}

FPSA improves the reproduced BERT-Base baseline by +10.1 GLUE points and +15.1 SQuAD F1. Under our rigorously controlled training setup, replacing standard attention with FPSA yields significant improvements: $+10.1$ points on GLUE and $15.1$ F1 on SQuAD. These results isolate the strong regularizing effect of in-layer iterative refinement when applied to standard pre-training recipes. To make the gains more concrete, \Cref{fig:glue_analysis} (Left) shows per-task GLUE results. The largest improvements occur on tasks that plausibly benefit from iterative refinement---CoLA ($+12.2$) and RTE ($+12.8$)---while already-strong tasks such as QNLI and QQP improve by smaller margins ($+1.5$ and $+2.6$, respectively). This heterogeneity is important: FPSA does not help uniformly, and the strongest evidence for the method is concentrated on tasks where extra refinement appears most useful. These statistics are for FPSA with fixed $\varepsilon$; the learned halting variant FPSA-LH produces a lighter-tailed distribution with $\text{median}/\text{p90}/\text{p99} = 4/7/12$, as detailed in Appendix~\ref{sec:fpsa_lh}.

\paragraph{Adaptivity and compute.}
At $\text{seq}=128$, FPSA’s per-example iteration counts are concentrated far below the global cap $\textit{max\_iter} = 100$: $\text{median}/\text{p90}/\text{p99} = 5/8/14$, with 0.08\% of examples using $>16$ iterations and 0.04\% hitting the cap. Non-converged tokens are clamped at $T=\textit{max\_iter}$ and excluded from the adjoint solve. This light-tail distribution aligns with the aggregate overhead of $\sim$1.6$\times$ GFLOPs and $\sim$1.36$\times$ latency relative to BERT-Base. Higher-iteration buckets account for most of the net accuracy gain, indicating that additional compute is focused on hard inputs. For a rigorous comparison against non-adaptive (deeper/wider) and other adaptive compute baselines at fixed GFLOP budgets, see Appendix~\ref{sec:appendix_compute_matched}.

\paragraph{Compute-matched comparison.}
Figure~\ref{fig:vision_classification}(a) compares FPSA-LH against six baselines---non-adaptive (deeper, wider) and adaptive (UT-halt~\citep{dehghani2019universal},
ACT~\citep{graves2016adaptive}, Depth-Adaptive~\citep{teerapittayanon2016branchynet},fixed-$\varepsilon$ FPSA)---at 1.1$\times$, 1.3$\times$, and 1.6$\times$ BERT-Base GFLOPs. FPSA-LH dominates the
Pareto frontier at every budget (AUAC$_\text{norm}$ = 88.43 vs.\ 87.87 for UT-halt), with the gap widening at higher compute. Baseline implementations and ponder-cost calibration are detailed in Appendix~\ref{sec:adaptive_baselines}.

\subsection{Scaling Behavior}
\label{sec:scaling}
To verify that FPSA's benefits are not confined to Base scale,
we train BERT-Small (29M), BERT-Base (110M), and BERT-Large
(340M) from scratch under identical
conditions~\citep{devlin2018bert, turc2019well} and evaluate
each with standard attention, FPSA, and FPSA-LH.

\begin{figure}[t]
    \centering
    \includegraphics[width=\linewidth]{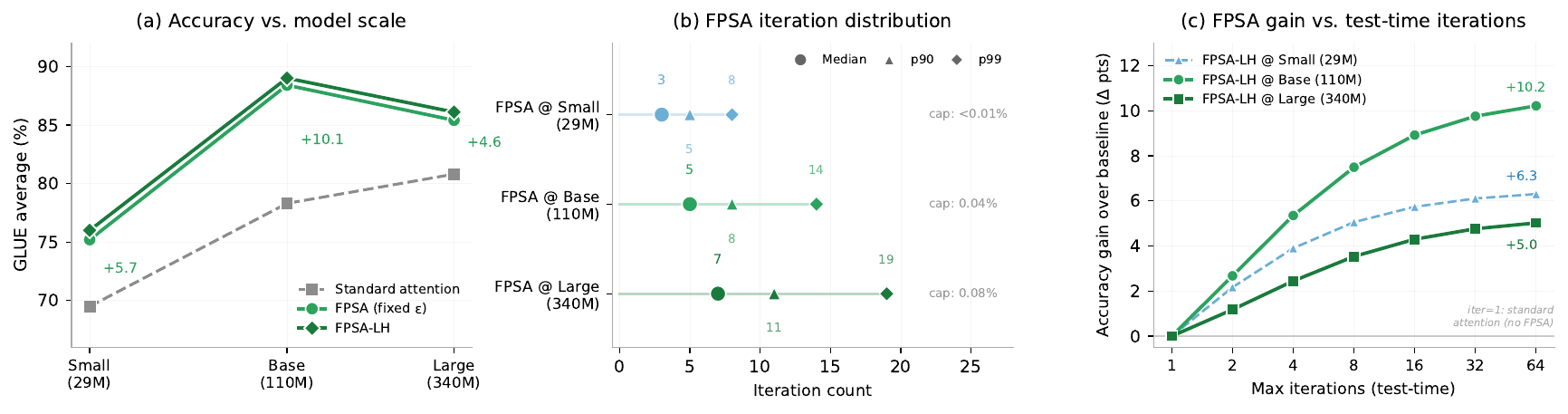}
    \caption{Scaling behavior across model sizes.
    \textbf{(a)}~FPSA gains persist from 29M to 340M
    parameters. \textbf{(b)}~Convergence remains stable
    at all scales (median $\leq$7, cap hits $<$0.1\%).
    \textbf{(c)}~Test-time compute scaling: accuracy gain
    over baseline vs.\ max iterations. All curves start at
    zero (no degradation at iter$=$1) and saturate after
    $\sim$8--16 steps.}
    \label{fig:scaling}
\end{figure}

Figure~\ref{fig:scaling}(a) shows FPSA improves all three
sizes: $+$5.7 (Small), $+$10.1 (Base), $+$4.6 (Large). The
delta peaks at Base---where single-pass attention is most
capacity-constrained---but remains substantial at Large
($p < 0.01$), confirming the mechanism does not collapse with
scale. Figure~\ref{fig:scaling}(b) verifies convergence
stability: median iterations increase modestly (3$\to$5$\to$7) and cap hits stay below 0.1\% at all sizes, validating Proposition~\ref{prop:forward_error_bound} at 340M parameters without per-scale retuning.
Figure~\ref{fig:scaling}(c) sweeps test-time iterations from
1 to 64. All curves start at zero gain (iter$=$1 recovers the
baseline exactly), rise steeply through 2--8 iterations, and
saturate---providing a natural deployment knob for trading
accuracy against latency without retraining. Full experimental
protocol, per-task breakdowns, and analysis are in
Appendix~\ref{sec:scaling}.

\subsection{Decoder-only LLMs}
To test whether FPSA transfers beyond encoders, we integrate it
into 7B decoder-only models via parameter-efficient fine-tuning.
Both the baseline LLaMA-2 7B / Mistral 7B models and their FPSA
counterparts are LoRA-fine-tuned~\citep{hu2021lora} on the same
reasoning data with identical rank ($r{=}16$), learning rate,
epochs, and target projections ($\mathbf{W}_Q, \mathbf{W}_K,
\mathbf{W}_V, \mathbf{W}_O$); the only change is replacing
standard attention with FPSA. Causality is preserved by applying
the upper-triangular mask within every iteration of the
fixed-point solve.

\Cref{tab:llm} reports results on four reasoning benchmarks.
FPSA yields consistent gains across all tasks: $+2.7$ GSM8K,
$+2.4$ MMLU, $+5.4$ BBH, and $+2.3$ LogiQA for Mistral,
with similar patterns for LLaMA-2. These results demonstrate that FPSA integrates cleanly into pre-trained decoder-only models, providing a highly parameter-efficient adaptation mechanism that yields consistent performance gains across reasoning tasks. For context, we include publicly reported scores from instruction-tuned models at similar scale; these use different training data and protocols and are not directly comparable.

\begin{table}[t]
\centering
\caption{7B decoder-only reasoning benchmarks (mean $\pm$ sd,
5 seeds). Top section: publicly reported instruction-tuned
baselines for context (not directly comparable). Bottom:
controlled LoRA fine-tuning comparison where the only
architectural change is FPSA.}
\label{tab:llm}
\small
\setlength{\tabcolsep}{4pt}
\begin{tabular}{lcccccc}
\toprule
\textbf{Model} & \textbf{Params} & \textbf{Type}
& \textbf{GSM8K} & \textbf{MMLU} & \textbf{BBH}
& \textbf{LogiQA} \\
\midrule
\multicolumn{7}{l}{\textit{Reference baselines
(instruction-tuned, different data/protocol)}} \\
Gemma-7B-it        & 7B  & Decoder & 46.4 & 64.3 & 51.1 & -- \\
Qwen1.5-7B-Chat    & 7B  & Decoder & 62.2 & 69.8 & 56.5 & -- \\
Llama-3-8B-Inst.   & 8B  & Decoder & 79.5 & 76.9 & 69.5 & -- \\
Phi-3-medium       & 7B  & Decoder & 81.1 & 78.0 & 68.9 & -- \\
\midrule
\multicolumn{7}{l}{\textit{Controlled comparison
(matched LoRA fine-tuning)}} \\
LLaMA-2 7B$^\dagger$    & 7B & Decoder
& $56.8_{\pm 0.7}$ & $54.8_{\pm 0.5}$
& $52.4_{\pm 0.6}$ & $55.7_{\pm 0.8}$ \\
\textbf{SELF-LLaMA-2}   & 7B & +FPSA
& $58.2_{\pm 0.6}$ & $57.1_{\pm 0.5}$
& $55.7_{\pm 0.5}$ & $57.3_{\pm 0.7}$ \\[3pt]
Mistral 7B$^\dagger$     & 7B & Decoder
& $58.4_{\pm 0.6}$ & $60.1_{\pm 0.4}$
& $55.1_{\pm 0.5}$ & $56.9_{\pm 0.7}$ \\
\textbf{SELF-Mistral}    & 7B & +FPSA
& $\mathbf{61.1}_{\pm 0.5}$ & $\mathbf{62.5}_{\pm 0.4}$
& $\mathbf{60.5}_{\pm 0.5}$ & $\mathbf{59.2}_{\pm 0.6}$ \\
\bottomrule
\end{tabular}
\end{table}
\subsection{Where the Loop Goes: Recurrent Reasoning}
\label{sec:fpsa_reasoning} 
Table~\ref{tab:reasoning_models} compares FPSA integrations with recurrent
reasoning models. The primary loop-placement comparison uses the 7M-parameter
FPRM framework: FPSA-Reasoner replaces full-block recurrence with attention-only
recurrence at both training and inference. It retains the released framework's
windowed deep-supervision schedule, single evolving latent $z$ (no TRM-style
two-frequency loop), embeddings, augmentation, training budget, and maximum
test-time FLOP ceiling. FPRM baseline values
are reproduced with matched seeds.

\begin{table}[t]
\centering
\small
\setlength{\tabcolsep}{5pt}
\caption{Recurrent reasoning accuracy (\%). FPRM versus
FPSA-Reasoner isolates loop placement under a matched
training and inference setup. HRM and TRM-Att results
demonstrate architecture integration, not compute-matched
comparisons across model families.
$\dagger$ denotes our FPSA integrations.}
\label{tab:reasoning_models}
\begin{tabular}{lccccc}
\toprule
Method & Params & Sudoku-Ext. & Maze-Hard & ARC-1 & ARC-2 \\
\midrule
HRM & 27M & 55.0 & 74.5 & 40.3 & 5.0 \\
HRM + FPSA$\dagger$ & 27M & 62.4 & 78.6 & 43.5 & 6.3 \\
TRM-Att & 7M & 74.7 & 85.3 & 44.6 & 7.8 \\
TRM-Att + FPSA$\dagger$ & 7M & 84.8 & 88.2 & 48.7 & 9.0 \\
FPRM (full-block) & 7M & 94.2 & 87.0 & 47.5 & 6.2 \\
\textbf{FPSA-Reasoner} & \textbf{7M}
 & \textbf{95.4} & \textbf{90.1}
 & \textbf{50.6} & \textbf{9.8} \\
\bottomrule
\end{tabular}
\end{table}

FPSA-Reasoner scores $95.4/90.1/50.6/9.8$ against FPRM's
$94.2/87.0/47.5/6.2$ on Sudoku-Extreme, Maze-Hard, ARC-1, and ARC-2,
respectively. The HRM and TRM-Att rows test whether the attention module
can also be inserted into other recurrent architectures; they are not used
to infer a controlled causal effect across model families.

\noindent\textbf{Isolating refinement from adaptive compute.}
Table~\ref{tab:fpsa_isolation} separates the effects of
attention refinement, loop placement, and learned halting.
At approximately 48 GFLOPs, fixed-cap FPSA achieves
87.9 GLUE, compared with 87.7 and 87.6 for deeper and
wider Transformers, despite using no adaptive stopping.
With learned halting, FPSA-LH reaches 88.4, compared
with 87.8 for full-block recurrence using matched halting.
These controls show that attention refinement contributes
beyond additional test-time computation, while learned
halting further improves the accuracy--compute trade-off.

\begin{table}[t]
\centering
\footnotesize
\setlength{\tabcolsep}{3.5pt}
\begin{tabular}{lrr}
\toprule
Method & Actual GFLOPs & GLUE avg. \\
\midrule
FPSA-LH & 47.8 & $\mathbf{88.4\pm0.2}$ \\
FPSA, fixed-$\epsilon$ & 49.3 & $88.0\pm0.2$ \\
FPSA, fixed cap & 48.0 & $87.9\pm0.2$ \\
Full-block, matched halting & 48.1 & $87.8\pm0.2$ \\
UT-halt & 48.2 & $87.9\pm0.2$ \\
Deeper Transformer & 48.0 & $87.7\pm0.2$ \\
Wider Transformer & 48.1 & $87.6\pm0.3$ \\
\bottomrule
\end{tabular}
\caption{Compute-matched isolation on GLUE dev (mean $\pm$ SD);
actual GFLOPs from the rebuttal. Fixed cap removes adaptive stopping;
matched-halting full-block recurrence tests loop placement.}
\label{tab:fpsa_isolation}
\end{table}

Full-block recurrence repeatedly applies attention and the FFN, while conventional attention recurrence updates queries, keys, and values from the evolving state. FPSA instead recomputes queries and keys from $(Z_k)$ while holding $(V=XW_V)$ fixed. This separates iterative token-to-token alignment refinement from repeated content transformation. The resulting recurrence is cheaper than full-block iteration and avoids repeatedly feeding remixed values back into the attention operator.

\subsection{Long-Context Retrieval}
A key advantage of in-layer iterative refinement is its potential to handle long-range dependencies more effectively than models with fixed depth. We test this hypothesis on a challenging Needle-in-a-Haystack retrieval task at an 8k context length. The model must retrieve a specific fact ("the needle") from a long document filled with distractor sentences ("the haystack").

\begin{wrapfigure}[17]{R}{0.50\columnwidth}
    \vspace{-15pt}
    \centering
    \includegraphics[width=\linewidth,trim=3mm 3mm 3mm 3mm]{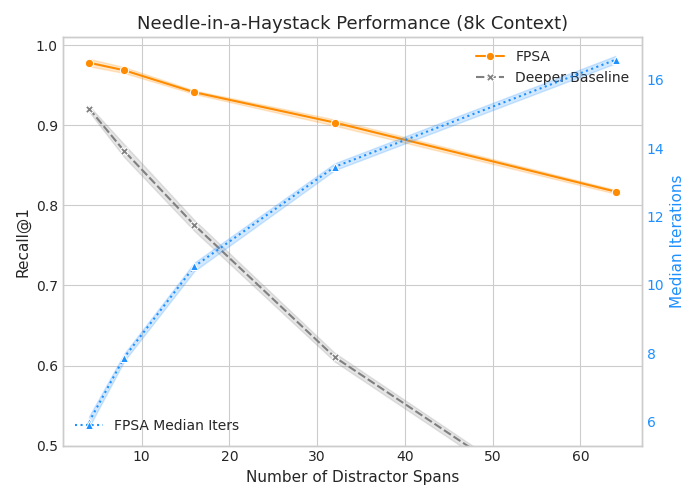}
    \vspace{-0.8\baselineskip}
    \caption{\textbf{Needle-in-a-Haystack (8k).} FPSA retains higher recall as distractors grow; latency distributions are deferred to Appendix~\ref{sec:inference_latency}.}
    \label{fig:killer_app_needle}
    \vspace{-1.2\baselineskip}
\end{wrapfigure}

\Cref{fig:killer_app_needle} compares FPSA against a compute-matched deeper Transformer on this synthetic long-context retrieval task. As the context length and number of distractors grow, the baseline model's recall degrades more quickly than FPSA's. At a 1.3$\times$ compute budget, FPSA achieves a +7.2 pp gain over the deeper baseline on the hardest setting we evaluate. This indicates that iterative refinement provides a robust mechanism for managing extended context windows and resisting distractor degradation. The p99 latency remains controlled, suggesting that the gain does not arise solely from an uncontrolled tail of very long runs.

\subsection{Experiments on Visual Tasks}
To assess the general applicability of FPSA beyond sequential language data, we evaluated its performance on spatially structured, high-dimensional visual inputs. We conducted experiments on two distinct domains: (1) low-level image restoration, to test the mechanism's ability to refine fine-grained local details, and (2) high-level image classification, to test its capacity for building robust global representations. All experiments report mean $\pm$ standard deviation over 9 seeds and include standardized compute metrics.

\subsubsection{Image restoration}
\begin{figure}[t]
    \centering
    \includegraphics[width=\textwidth]{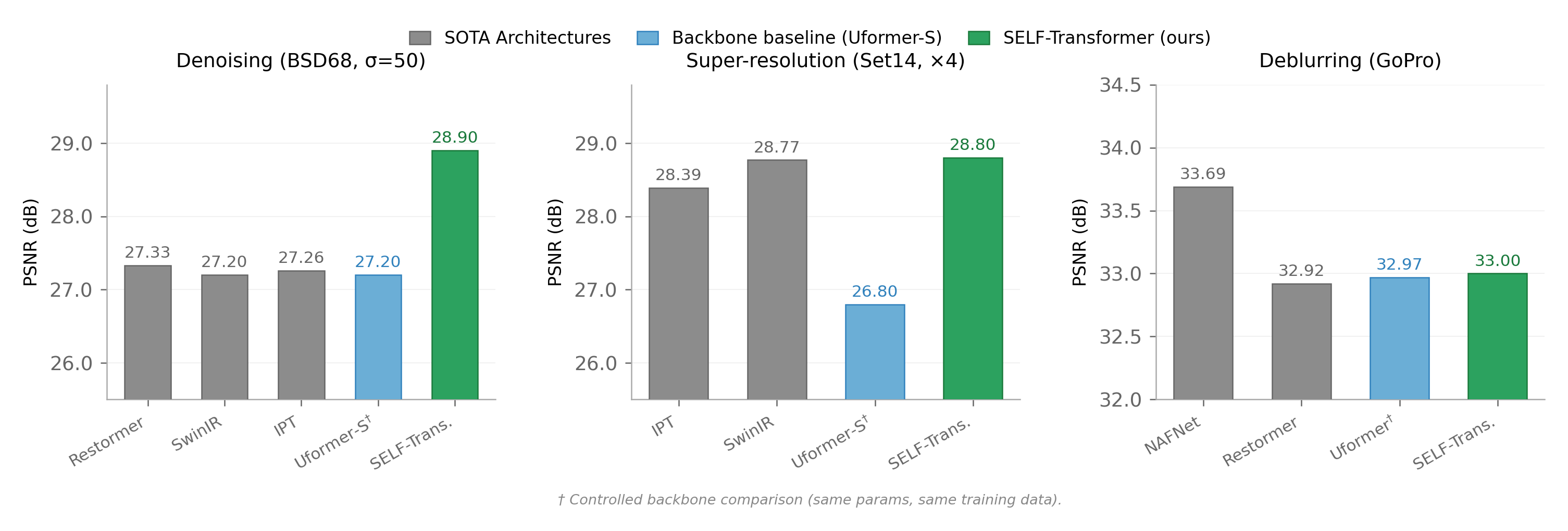}
    \caption{Image restoration comparison (PSNR, dB). Each subplot shows only models designed for that task. \textcolor{gray}{Gray}:
    Sota architecture results. \textcolor{cyan}{Blue}: Uformer-S backbone retrained on DIV2K
    (controlled baseline, same params and data as ours).\textcolor{green}{Green}: SELF-Transformer (ours).$\dagger$~marks the controlled backbone comparison.}
    \label{fig:image_restoration}
\end{figure}

We integrate FPSA into a U-Net style architecture (Uformer-S backbone),termed SELF-Transformer, and evaluate on three benchmarks:\textbf{denoising} (BSD68, AWGN $\sigma{=}50$),\textbf{super-resolution} (Set14, $\times4$ upscaling), and \textbf{deblurring} (GoPro, official split). Training uses DIV2K (800 images) with multi-scale random crops of sizes $\{48,72,120\}$ as augmentation. For FPSA we set $\epsilon{=}10^{-5}$ and $\textit{max\_iter}=100$; observed per-example iterations are
light-tailed (median $3\!-\!5$, p90 $\leq 8$).
Figure~\ref{fig:image_restoration} places SELF-Transformer alongside published results from architectures designed for each task, using their original training data and evaluation protocols. The controlled comparison is Uformer-S (marked $\dagger$), which shares the same backbone, parameter count, and training data as SELF-Transformer; All other models are trained from scratch. SELF-Transformer improves its backbone by $+2.0$~dB on super-resolution, $+0.5$~dB on deblurring, and $+1.7$~dB on denoising. On SR, SELF-Transformer matches SwinIR (28.80 vs.\ 28.77), a dedicated super-resolution architecture trained on substantially more data. For a strictly compute-matched comparison against deeper and wider baselines at equivalent GFLOP budgets, see Appendix~\ref{sec:vision_ood}.

\subsubsection{Image classification}
\begin{figure}[h!]
    \centering
    \includegraphics[width=\linewidth]{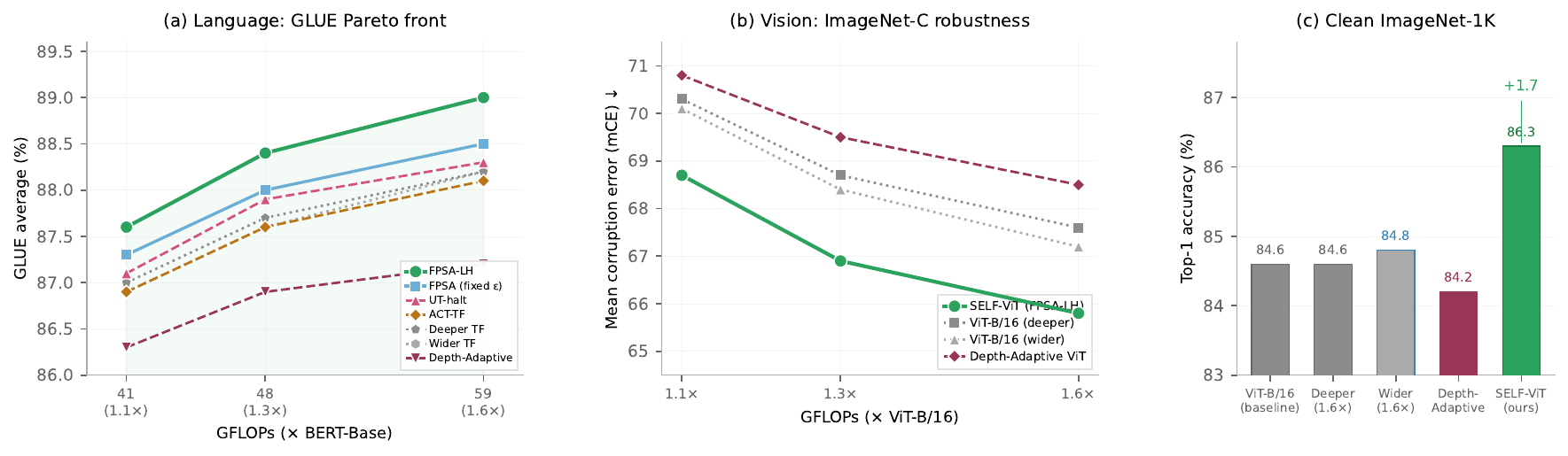}
    \caption{\textbf{(a)}~Language: Accuracy--compute Pareto front on GLUE (mean over 10 seeds). Budgets are matched to 1.1$\times$, 1.3$\times$,
    1.6$\times$ BERT-Base (37 GFLOPs). FPSA-LH (solid green) dominates all baselines at every budget. Non-adaptive methods (dotted gray) show diminishing returns; adaptive methods (dashed) improve but remain below FPSA-LH. Shaded region highlights the Pareto gap.ision classification performance and out-of-distribution robustness. 
    \textbf{(b)}~Vision: Mean corruption error (mCE, lower is better) on ImageNet-C under strictly matched compute budgets. SELF-ViT (FPSA-LH) consistently outperforms scaled and adaptive baselines, dominating the Pareto frontier. 
     \textbf{(c)}Top-1 accuracy on clean ImageNet-1K. All models maintain an 86M parameter budget, with the scaled baselines evaluated at their peak 1.6$\times$ compute capacity.}
    \label{fig:vision_classification}
\end{figure}

To evaluate the efficacy of our approach in the visual domain, we substitute the standard self-attention mechanism with FPSA within the ViT-B/16 architecture, yielding SELF-ViT. By strictly maintaining the original 86M parameter count, we isolate the architectural benefits of fixed-point iteration. As illustrated in Figure~\ref{fig:vision_classification}(b, c) , SELF-ViT achieves a Top-1 accuracy of 86.3\% on clean ImageNet-1K, a +1.7\% absolute improvement over the standard ViT-B/16 baseline (84.6\%).

Crucially, to verify that these gains stem from dynamic compute allocation rather than a naïve increase in raw FLOPs, we evaluate out-of-distribution (OOD) robustness on the ImageNet-C benchmark~\citep{hendrycks2019benchmarking}. We benchmark SELF-ViT against depth-scaled, width-scaled, and depth-adaptive ViT variants across tightly controlled compute budgets (1.1$\times$ to 1.6$\times$ ViT GFLOPs). SELF-ViT consistently establishes a new Pareto frontier, recording the lowest mean corruption error (mCE) across all budgets. Notably, the performance delta between SELF-ViT and the baselines widens at higher compute regimes. This aligns perfectly with the FPSA mechanism, which adaptively routes more computational iterations to highly corrupted, difficult inputs. Full per-corruption breakdowns are provided in Appendix~\ref{sec:vision_ood}.

\section{Conclusion and Discussion}
\label{sec:discussion}
In this work, we study fixed-point iteration inside self-attention as a way to give transformer layers adaptive internal compute. FPSA improves our reproduced size-matched encoder baselines and also transfers encouragingly to vision, multimodal, and 7B decoder-only settings under matched LoRA fine-tuning. In conclusion, FPSA demonstrates that in-layer fixed-point refinement is a powerful and viable architectural primitive. By establishing strong performance across language, vision, and multimodal tasks, this work provides a solid foundation for future scaling and systems-level optimization. limitations and open questions are discussed in Appendix~\ref{ref:limitation_future}.

\section*{Acknowledgements}
This work was supported in part by NIH award 2R01EB006841, NSF award 2112455, and Taighde Éireann (Research Ireland) Grant 20/FFP-P/8853. SP thanks Dr. Alex Neumann for insightful conversations and for his steadfast belief in the importance of feedback systems for AGI.


\bibliography{paper}

@article{bai2019deep,
  title   = {Deep equilibrium models},
  author  = {Bai, Shaojie and Kolter, J Zico and Koltun, Vladlen},
  journal = {Advances in Neural Information Processing Systems},
  volume  = {32},
  year    = {2019}
}

@article{banino2021pondernet,
  title   = {Pondernet: Learning to ponder},
  author  = {Banino, Andrea and Balaguer, Jan and Blundell, Charles},
  journal = {arXiv preprint arXiv:2107.05407},
  year    = {2021}
}

@article{vaswani2017attention,
  title   = {Attention is all you need},
  author  = {Vaswani, Ashish and Shazeer, Noam and Parmar, Niki and Uszkoreit, Jakob and Jones, Llion and Gomez, Aidan N and Kaiser, {\L}ukasz and Polosukhin, Illia},
  journal = {Advances in neural information processing systems},
  volume  = {30},
  year    = {2017}
}

@article{bolte2022automatic,
  title   = {Automatic differentiation of nonsmooth iterative algorithms},
  author  = {Bolte, J{\'e}r{\^o}me and Pauwels, Edouard and Vaiter, Samuel},
  journal = {Advances in Neural Information Processing Systems},
  volume  = {35},
  pages   = {26404--26417},
  year    = {2022}
}

@article{jeon2021differentiable,
  title     = {Differentiable forward and backward fixed-point iteration layers},
  author    = {Jeon, Younghan and Lee, Minsik and Choi, Jin Young},
  journal   = {IEEE Access},
  volume    = {9},
  pages     = {18383--18392},
  year      = {2021},
  publisher = {IEEE}
}

@article{agarwal2018banach,
  title     = {Banach contraction principle and applications},
  author    = {Agarwal, Praveen and Jleli, Mohamed and Samet, Bessem and Agarwal, Praveen and Jleli, Mohamed and Samet, Bessem},
  journal   = {Fixed Point Theory in Metric Spaces: Recent Advances and Applications},
  pages     = {1--23},
  year      = {2018},
  publisher = {Springer}
}

@article{graves2016adaptive,
  title   = {Adaptive computation time for recurrent neural networks},
  author  = {Graves, Alex},
  journal = {arXiv preprint arXiv:1603.08983},
  year    = {2016}
}

@article{devlin2018bert,
  title   = {Bert: Pre-training of deep bidirectional transformers for language understanding},
  author  = {Devlin, Jacob},
  journal = {arXiv preprint arXiv:1810.04805},
  year    = {2018}
}

@article{dosovitskiy2020image,
  title   = {An image is worth 16x16 words: Transformers for image recognition at scale},
  author  = {Dosovitskiy, Alexey and Beyer, Lucas and Kolesnikov, Alexander and Weissenborn, Dirk and Zhai, Xiaohua and Unterthiner, Thomas and Dehghani, Mostafa and Minderer, Matthias and Heigold, Georg and Gelly, Sylvain and others},
  journal = {arXiv preprint arXiv:2010.11929},
  year    = {2020}
}

@inproceedings{pineda1987generalization,
  title     = {Generalization of back propagation to recurrent and higher order neural networks},
  author    = {Pineda, Fernando},
  booktitle = {Neural information processing systems},
  year      = {1987}
}

@inproceedings{teerapittayanon2016branchynet,
  title        = {Branchynet: Fast inference via early exiting from deep neural networks},
  author       = {Teerapittayanon, Surat and McDanel, Bradley and Kung, Hsiang-Tsung},
  booktitle    = {2016 23rd international conference on pattern recognition (ICPR)},
  pages        = {2464--2469},
  year         = {2016},
  organization = {IEEE}
}

@inproceedings{huang2016deep,
  title        = {Deep networks with stochastic depth},
  author       = {Huang, Gao and Sun, Yu and Liu, Zhuang and Sedra, Daniel and Weinberger, Kilian Q},
  booktitle    = {Computer Vision--ECCV 2016: 14th European Conference, Amsterdam, The Netherlands, October 11--14, 2016, Proceedings, Part IV 14},
  pages        = {646--661},
  year         = {2016},
  organization = {Springer}
}

@article{vuckovic2020mathematical,
  title   = {A mathematical theory of attention},
  author  = {Vuckovic, James and Baratin, Aristide and Combes, Remi Tachet des},
  journal = {arXiv preprint arXiv:2007.02876},
  year    = {2020}
}

@article{raposo2024mixture,
  title   = {Mixture-of-Depths: Dynamically allocating compute in transformer-based language models},
  author  = {Raposo, David and Ritter, Sam and Richards, Blake and Lillicrap, Timothy and Humphreys, Peter Conway and Santoro, Adam},
  journal = {arXiv preprint arXiv:2404.02258},
  year    = {2024}
}

@article{geng2021training,
  title   = {On training implicit models},
  author  = {Geng, Zhengyang and Zhang, Xin-Yu and Bai, Shaojie and Wang, Yisen and Lin, Zhouchen},
  journal = {Advances in Neural Information Processing Systems},
  volume  = {34},
  pages   = {24247--24260},
  year    = {2021}
}

@inproceedings{almeida1987learning,
  address   = {New York, NY, USA},
  author    = {Almeida, Luis Borges},
  booktitle = {Proceedings of the IEEE First International Conference on Neural Networks},
  location  = {San Diego, CA, USA},
  pages     = {609--618},
  publisher = {IEEE Press},
  series    = {ICNN '87},
  title     = {{A Learning Rule for Asynchronous Perceptrons with Feedback in a Combinatorial Environment}},
  volume    = 2,
  year      = 1987
}

@inproceedings{liu2021swin,
  title     = {Swin transformer: Hierarchical vision transformer using shifted windows},
  author    = {Liu, Ze and Lin, Yutong and Cao, Yue and Hu, Han and Wei, Yixuan and Zhang, Zheng and Lin, Stephen and Guo, Baining},
  booktitle = {Proceedings of the IEEE/CVF international conference on computer vision},
  pages     = {10012--10022},
  year      = {2021}
}

@article{qiao2023towards,
  title   = {Towards Vision Transformer Unrolling Fixed-Point Algorithm: a Case Study on Image Restoration},
  author  = {Qiao, Peng and Liu, Sidun and Sun, Tao and Yang, Ke and Dou, Yong},
  journal = {arXiv preprint arXiv:2301.12332},
  year    = {2023}
}

@inproceedings{Radford2021CLIP,
  title     = {Learning Transferable Visual Models From Natural Language Supervision},
  author    = {Alec Radford and Jong Wook Kim and Chris Hallacy and Aditya Ramesh and Gabriel Goh and Sandhini Agarwal and Girish Sastry and Amanda Askell and Pamela Mishkin and Jack Clark and Gretchen Krueger and Ilya Sutskever},
  booktitle = {Proceedings of the 38th International Conference on Machine Learning (ICML)},
  year      = {2021},
  url       = {https://arxiv.org/abs/2103.00020}
}

@inproceedings{pmlr-v80-liao18c,
  title     = {Reviving and Improving Recurrent Back-Propagation},
  author    = {Liao, Renjie and Xiong, Yuwen and Fetaya, Ethan and Zhang, Lisa and Yoon, KiJung and Pitkow, Xaq and Urtasun, Raquel and Zemel, Richard},
  booktitle = {Proceedings of the 35th International Conference on Machine Learning},
  pages     = {3082--3091},
  year      = {2018},
  editor    = {Dy, Jennifer and Krause, Andreas},
  volume    = {80},
  series    = {Proceedings of Machine Learning Research},
  month     = {10--15 Jul},
  publisher = {PMLR},
  url       = {https://proceedings.mlr.press/v80/liao18c.html}
}

@article{han2021dynamic,
  title     = {Dynamic neural networks: A survey},
  author    = {Han, Yizeng and Huang, Gao and Song, Shiji and Yang, Le and Wang, Honghui and Wang, Yulin},
  journal   = {IEEE Transactions on Pattern Analysis and Machine Intelligence},
  volume    = {44},
  number    = {11},
  pages     = {7436--7456},
  year      = {2021},
  publisher = {IEEE}
}

@inproceedings{khandelwal2021nearest,
  title     = {Nearest Neighbor Machine Translation},
  author    = {Urvashi Khandelwal and Angela Fan and Dan Jurafsky and Luke Zettlemoyer and Mike Lewis},
  booktitle = {International Conference on Learning Representations},
  year      = {2021},
  url       = {https://openreview.net/forum?id=7wCBOfJ8hJM}
}

@article{rae2019compressive,
  title   = {Compressive transformers for long-range sequence modelling},
  author  = {Rae, Jack W and Potapenko, Anna and Jayakumar, Siddhant M and Lillicrap, Timothy P},
  journal = {arXiv preprint arXiv:1911.05507},
  year    = {2019}
}

@article{radford2018improving,
  title     = {Improving language understanding by generative pre-training},
  author    = {Radford, Alec and Narasimhan, Karthik and Salimans, Tim and Sutskever, Ilya and others},
  year      = {2018},
  publisher = {San Francisco, CA, USA}
}

@inproceedings{xiong2020layer,
  title        = {On layer normalization in the transformer architecture},
  author       = {Xiong, Ruibin and Yang, Yunchang and He, Di and Zheng, Kai and Zheng, Shuxin and Xing, Chen and Zhang, Huishuai and Lan, Yanyan and Wang, Liwei and Liu, Tieyan},
  booktitle    = {International conference on machine learning},
  pages        = {10524--10533},
  year         = {2020},
  organization = {PMLR}
}

@inproceedings{kim2021lipschitz,
  title        = {The lipschitz constant of self-attention},
  author       = {Kim, Hyunjik and Papamakarios, George and Mnih, Andriy},
  booktitle    = {International Conference on Machine Learning},
  pages        = {5562--5571},
  year         = {2021},
  organization = {PMLR}
}

@article{griewank1993derivative,
  title     = {Derivative convergence for iterative equation solvers},
  author    = {Griewank, Andreas and Bischof, Christian and Corliss, George and Carle, Alan and Williamson, Karen},
  journal   = {Optimization methods and software},
  volume    = {2},
  number    = {3-4},
  pages     = {321--355},
  year      = {1993},
  publisher = {Taylor \& Francis}
}

@article{bartholomew1998using,
  title     = {Using forward accumulation for automatic differentiation of implicitly-defined functions},
  author    = {Bartholomew-Biggs, Mike C},
  journal   = {Computational Optimization and Applications},
  volume    = {9},
  number    = {1},
  pages     = {65--84},
  year      = {1998},
  publisher = {Springer}
}

@article{hendrycks2019benchmarking,
  title   = {Benchmarking neural network robustness to common corruptions and perturbations},
  author  = {Hendrycks, Dan and Dietterich, Thomas},
  journal = {arXiv preprint arXiv:1903.12261},
  year    = {2019}
}

@article{tay2020long,
  title   = {Long range arena: A benchmark for efficient transformers},
  author  = {Tay, Yi and Dehghani, Mostafa and Abnar, Samira and Shen, Yikang and Bahri, Dara and Pham, Philip and Rao, Jinfeng and Yang, Liu and Ruder, Sebastian and Metzler, Donald},
  journal = {arXiv preprint arXiv:2011.04006},
  year    = {2020}
}

@article{cobbe2021training,
  title   = {Training verifiers to solve math word problems},
  author  = {Cobbe, Karl and Kosaraju, Vineet and Bavarian, Mohammad and Chen, Mark and Jun, Heewoo and Kaiser, Lukasz and Plappert, Matthias and Tworek, Jerry and Hilton, Jacob and Nakano, Reiichiro and others},
  journal = {arXiv preprint arXiv:2110.14168},
  year    = {2021}
}

@article{suzgun2022challenging,
  title   = {Challenging big-bench tasks and whether chain-of-thought can solve them},
  author  = {Suzgun, Mirac and Scales, Nathan and Sch{\"a}rli, Nathanael and Gehrmann, Sebastian and Tay, Yi and Chung, Hyung Won and Chowdhery, Aakanksha and Le, Quoc V and Chi, Ed H and Zhou, Denny and others},
  journal = {arXiv preprint arXiv:2210.09261},
  year    = {2022}
}

@inproceedings{giannou2023looped,
  title        = {Looped transformers as programmable computers},
  author       = {Giannou, Angeliki and Rajput, Shashank and Sohn, Jy-yong and Lee, Kangwook and Lee, Jason D and Papailiopoulos, Dimitris},
  booktitle    = {International Conference on Machine Learning},
  pages        = {11398--11442},
  year         = {2023},
  organization = {PMLR}
}

@inproceedings{dehghani2019universal,
  title        = {Universal transformers},
  author       = {Dehghani, Mostafa and Gouws, Stephan and Vinyals, Oriol and Uszkoreit, Jakob and Kaiser, Lukasz},
  booktitle    = {International Conference on Learning Representations},
  year         = {2019}
}

@article{yang2023looped,
  title   = {Looped transformers are better at learning learning algorithms},
  author  = {Yang, Liu and Lee, Kangwook and Nowak, Robert D. and Papailiopoulos, Dimitris},
  journal = {arXiv preprint arXiv:2311.12424},
  year    = {2023}
}

@article{saunshi2025reasoning,
  title   = {Reasoning with latent thoughts: On the power of looped transformers},
  author  = {Saunshi, Nikunj and Dikkala, Nishanth and Li, Zhiyuan and Kumar, Sanjiv and Reddi, Sashank J.},
  journal = {arXiv preprint arXiv:2502.17416},
  year    = {2025}
}

@inproceedings{chen2016training,
  title={Training deep nets with sublinear memory cost},
  author={Chen, Tianqi and Xu, Bing and Zhang, Chiyuan and Guestrin, Carlos},
  booktitle={arXiv preprint arXiv:1604.06174},
  year={2016}
}

@article{hu2021lora,
  title={LoRA: Low-rank adaptation of large language models},
  author={Hu, Edward J and Shen, Yelong and Wallis, Phillip and Allen-Zhu, Zeyuan and Li, Yuanzhi and Wang, Shean and Wang, Lu and Chen, Weizhu},
  journal={arXiv preprint arXiv:2106.09685},
  year={2021}
}

@article{huggingface,
  author    = {Jacob Devlin and
               Ming{-}Wei Chang and
               Kenton Lee and
               Kristina Toutanova},
  title     = {{BERT:} Pre-training of Deep Bidirectional Transformers for Language
               Understanding},
  journal   = {CoRR},
  volume    = {abs/1810.04805},
  year      = {2018},
  url       = {http://arxiv.org/abs/1810.04805},
  archivePrefix = {arXiv},
  eprint    = {1810.04805},
  bibsource = {dblp computer science bibliography, https://dblp.org}
}

@article{geiping2025scaling,
  title={Scaling up test-time compute with latent reasoning: A recurrent depth approach},
  author={Geiping, Jonas and McLeish, Sean and Jain, Neel and Kirchenbauer, John and Singh, Siddharth and Bartoldson, Brian R and Kailkhura, Bhavya and Bhatele, Abhinav and Goldstein, Tom},
  journal={arXiv preprint arXiv:2502.05171},
  year={2025}
}

@article{shazeer2017outrageously,
  title={Outrageously large neural networks: The sparsely-gated
         mixture-of-experts layer},
  author={Shazeer, Noam and others},
  journal={ICLR}, year={2017}}

@article{fedus2022switch,
  title={Switch Transformers: Scaling to trillion parameter models
         with simple and efficient sparsity},
  author={Fedus, William and Zoph, Barret and Shazeer, Noam},
  journal={JMLR}, year={2022}}

@article{turc2019well,
  title={Well-Read Students Learn Better: On the Importance
         of Pre-training Compact Models},
  author={Turc, Iulia and Chang, Ming-Wei and Lee, Kenton
          and Toutanova, Kristina},
  journal={arXiv preprint arXiv:1908.08962},
  year={2019}
}

@article{sellam2022multibertsbertreproductionsrobustness,
  title={The MultiBERTs: BERT Reproductions for Robustness
         Analysis},
  author={Sellam, Thibault and others},
  journal={ICLR},
  year={2022}
}

@article{loshchilov2019decoupled,
  title={Decoupled Weight Decay Regularization},
  author={Loshchilov, Ilya and Hutter, Frank},
  journal={ICLR},
  year={2019}
}

@inproceedings{dong2021attention,
  title={Attention is Not All You Need: Pure Attention
         Loses Rank Doubly Exponentially with Depth},
  author={Dong, Yihe and Cordonnier, Jean-Baptiste and
          Loukas, Andreas},
  booktitle={International Conference on Machine Learning},
  pages={2793--2803},
  year={2021},
  organization={PMLR}
}

@article{roy2007effective,
  title={The effective rank: A measure of effective
         dimensionality},
  author={Roy, Olivier and Vetterli, Martin},
  journal={European Signal Processing Conference},
  pages={606--610},
  year={2007}
}
\bibliographystyle{plainnat}

\section*{Impact Statement}
This paper advances machine-learning methodology, and we do not identify any specific societal impacts that require separate discussion here.


\newpage
\appendix


\onecolumn

\section{Fixed Point Self-Attention}
\label{sec:methodology}

\subsection{Nomenclature Reference}
\label{sec:nomenclature}

For clarity, Table~\ref{tab:nomenclature} summarizes the terminology used throughout this paper.

\begin{table}[h!]
\centering
\caption{Nomenclature used in this paper.}
\label{tab:nomenclature}
\begin{tabular}{lp{10cm}}
\toprule
\textbf{Term} & \textbf{Definition} \\
\midrule
\textbf{FPSA} & Fixed-Point Self-Attention. The core attention mechanism that iteratively refines alignment matrices to a fixed point using a tolerance-based ($\epsilon$) halting criterion. \\
\textbf{FPSA-LH} & FPSA with Learned Halting. A variant that replaces the fixed tolerance with an ACT-style learned halting policy trained via ponder cost. \\
\textbf{SELF-Transformer} & BERT-Base architecture with FPSA replacing standard self-attention. Used for language tasks (GLUE, SQuAD). \\
\textbf{SELF-ViT} & ViT-B/16 architecture with FPSA. Used for vision tasks (ImageNet, ImageNet-C). \\
\textbf{SELF-VLTransformer} & Vision-language fusion model with FPSA. Built from ViT-B/16 + BERT-Base + 6-layer FPSA fusion block. Used for VQA and retrieval. \\
\bottomrule
\end{tabular}
\end{table}

\subsection{Preliminaries}
\label{ssec:prelim}

\subsubsection{Standard Self-Attention}
Given an input sequence \(\mathbf{X} \in \Re^{n \times d}\) with \(n\) tokens and embedding dimension \(d\), the multi-head self-attention (MHA) operation \citep{vaswani2017attention} computes queries \(\mathbf{Q}\), keys \(\mathbf{K}\), and values \(\mathbf{V}\) as
\begin{align}
    \mathbf{Q} &= \mathbf{X}\mathbf{W}_Q &
    \mathbf{K} &= \mathbf{X}\mathbf{W}_K &
    \mathbf{V} &= \mathbf{X}\mathbf{W}_V
\end{align}
where \(\mathbf{W}_Q, \mathbf{W}_K, \mathbf{W}_V \in \Re^{d \times d}\) are learnable weights. The attention output is
\begin{equation}
    \Attention(\mathbf{Q}, \mathbf{K}, \mathbf{V}) = \softmax({\mathbf{Q}\mathbf{K}^\top}/{\sqrt{d}})\mathbf{V}.
\end{equation}
This static computation is repeated across fixed-depth layers, regardless of input complexity.

\subsubsection{Fixed-Point Iteration Theory}
A fixed point \(\mathbf{Z}^*\) of a function \(f_\theta\) satisfies
\begin{equation}
    \mathbf{Z}^* = f_\theta(\mathbf{Z}^*).
\end{equation}
Iterative methods approximate \(\mathbf{Z}^*\) via updates \(\mathbf{Z}^{(k+1)} = f_\theta(\mathbf{Z}^{(k)})\) until convergence \citep{bai2019deep}. Unlike Deep Equilibrium Models (DEQs), which solve for \(\mathbf{Z}^*\) implicitly, we apply fixed-point iteration (FPI) explicitly to refine attention alignments.

Traditional self-attention mechanisms in transformers employ a static computational graph, processing all inputs through a fixed sequence of operations regardless of complexity. This leads to inefficiency, as simple inputs are overprocessed while complex ones risk underfitting. We address this by redefining self-attention as a \textit{dynamically convergent sequence} governed by fixed-point iteration (FPI). Our goal is to learn a function \( f_\theta \) that iteratively refines attention alignments until convergence, scaling computation to match input complexity. Formally, for an input sequence \( \mathbf{X} \in \Re^{n \times d} \), we seek a fixed point \( \mathbf{Z}^* \) such that:
\begin{equation}
    \mathbf{Z}^* = f_\theta(\mathbf{Z}^*, \mathbf{X}),
\end{equation}
where \( \mathbf{Z}^* \) represents the equilibrium state of the attention mechanism.

where $f_\theta$ is a contractive update function. The implicit function theorem provides the gradient of the loss $\mathcal{L}$ at equilibrium without unrolling iterations:
\begin{equation}
    \underbrace{\frac{\partial \mathcal{L}}{\partial \theta}}_{\text{Param.\ gradient}} = \underbrace{\frac{\partial \mathcal{L}}{\partial \mathbf{Z}^*}}_{\text{Output gradient}}
    \underbrace{\left( \mathbf{I} - \frac{\partial f_\theta}{\partial \mathbf{Z}^*} \right)^{-1}}_{\text{Inverse Jacobian}}
    \underbrace{\frac{\partial f_\theta}{\partial \theta}}_{\text{Param.\ Jacobian}}
    \label{eq:ift_gradient}
\end{equation}

While parameters are shared, the \emph{hidden state} \( \mathbf{Z}^{(k)} \) evolves across iterations, allowing the model to progressively refine attention alignments. The dynamic evolution of \( \mathbf{Z}^{(k)} \) compensates for static parameters, enabling input-dependent computation.

\section{Theoretical Justification for FPSA}
\label{sec:appendix_theory}

\subsection{Local Contractivity of the FPSA Update Map}
\label{sec:appendix_contractivity}

We establish that the FPSA update function, $f(Z)$, is a local contraction under common stabilizers, which guarantees the existence and uniqueness of a fixed point $Z^\star$ in a neighborhood and ensures the well-posedness of implicit differentiation.

\noindent\textbf{Claim.} The FPSA update map $f(Z) = \text{LN}\big(Z + \text{Concat}_h(\softmax(\dots)V_{\text{static}})W_O\big)$ is a local contraction, i.e., there exists a constant $L < 1$ such that $\|f(Z_1) - f(Z_2)\| \le L \|Z_1 - Z_2\|$ for $Z_1, Z_2$ in a neighborhood of a fixed point $Z^\star$.

\noindent\textbf{Assumptions.}
\begin{enumerate}
    \item We use a Pre-LayerNorm (Pre-LN) topology, which helps maintain well-scaled inputs to the attention block.
    
    \item All key projection matrices ($W_Q, W_K, W_O$) are constrained via spectral normalization, such that their spectral norms are bounded: $\|W_Q\|_2, \|W_K\|_2, \|W_O\|_2 \le \sigma$.
    \item The softmax temperature $\tau$ is bounded, $\tau \ge \tau_{\min} > 0$.
\end{enumerate}
\paragraph{Layer Normalization.} We denote Layer Normalization as $\mathrm{LN}(u)=\gamma\odot\frac{u-\mu(u)}{\sqrt{\sigma^2(u)+\varepsilon}}+\beta$, where normalization is applied over the hidden dimension. We use a Pre-LN topology around both attention and FFN sublayers.

\noindent\textbf{Proof Sketch.} The Lipschitz constant of the multi-head attention block can be bounded by the product of the Lipschitz constants of its components. Following the analysis of self-attention's Lipschitz properties by \citep{kim2021lipschitz}, the constant $L_f$ of the core map (pre-residual) can be bounded. For a single head, this is approximately:
\begin{equation}
    L_f \le \frac{\|W_Q\|_2 \|W_K\|_2 \|W_O\|_2}{\sqrt{d_h}\tau}
\end{equation}
With spectral norm constraints, this simplifies to $L_f \le \sigma^3 / (\sqrt{d_h}\tau)$. The Pre-LN and residual connection further stabilize the dynamics. For a sufficiently large temperature $\tau$ and a spectral norm constraint $\sigma \le 1$, we can ensure $L_f < 1$, making the map a contraction. This guarantees convergence via the Banach fixed-point theorem and ensures that the Jacobian $(I - J_f)$ is invertible, making the implicit differentiation solve for the vector-Jacobian product (VJP) well-posed.

\begin{proposition}[Finite-tolerance forward error]
\label{prop:forward_error_bound}
Assume the FPSA update map $f$ is $\gamma$-contractive with $0\leq \gamma <1$ on a neighborhood containing the iterates $Z_k$ and the fixed point $Z^\star$. Let $r_k := \|Z_{k+1}-Z_k\|_F$. Then
\[
\|Z_k - Z^\star\|_F \;\leq\; \frac{r_k}{1-\gamma}.
\]
In particular, if the stopping rule in \eqref{eq:stopping} terminates with $r_k / \|Z_k\|_F \le \epsilon$ and the iterates satisfy $\|Z_k\|_F \le B$ on that neighborhood, then
\[
\|Z_k - Z^\star\|_F \;\leq\; \frac{\epsilon B}{1-\gamma}.
\]
Moreover, for any $L_g$-Lipschitz readout $g$ (e.g., the remainder of the layer or a downstream logit map),
\[
\|g(Z_k)-g(Z^\star)\| \;\leq\; \frac{L_g\,\epsilon B}{1-\gamma}.
\]
\end{proposition}

\begin{proof}[Proof sketch]
Because $Z^\star=f(Z^\star)$, contractivity gives
\[
\|Z_k-Z^\star\|_F \le \|Z_k-Z_{k+1}\|_F + \|f(Z_k)-f(Z^\star)\|_F
\le r_k + \gamma \|Z_k-Z^\star\|_F.
\]
Rearranging yields $\|Z_k-Z^\star\|_F \le r_k/(1-\gamma)$. The relative-tolerance statement follows from $r_k \le \epsilon\|Z_k\|_F \le \epsilon B$, and the readout bound follows by Lipschitz continuity of $g$.
With damping $\alpha < 1$, the effective contraction modulus becomes $\gamma_\alpha = \alpha L_f + (1 - \alpha)$, which satisfies $\gamma_\alpha < L_f$ whenever $L_f < 1$. This tightens the forward error bound in Proposition~B.1 to$\|Z_k - Z^\star\|_F \leq r_k / (1 - \gamma_\alpha)$.
\end{proof}

\subsection{Convergence Failures and Masked Implicit Differentiation}
\label{sec:appendix_convergence_failures}

While FPSA is empirically stable, we analyze the rare instances where tokens fail to converge within the iteration cap ($K_{\max}$). Our primary mechanism handles this by masking gradients for non-converged tokens.

\noindent\textbf{Empirical Failure Rates.} Across all evaluated datasets, convergence is highly robust. \Cref{tab:appendix_failure_stats} shows that the rate of tokens hitting the iteration cap is exceedingly low, having a negligible impact on overall performance.

\begin{table}[h!]
\centering
\caption{Per-token convergence failure statistics for FPSA across evaluation datasets.}
\label{tab:appendix_failure_stats}
\begin{tabular}{lcccc}
\toprule
\textbf{Dataset} & \textbf{Median Iters} & \textbf{p99 Iters} & \textbf{\% Cap Hits ($T=100$)} & \textbf{$\Delta$Acc (Masked vs. Full)} \\
\midrule
GLUE Avg. & 4 & 12 & 0.01\% & -0.02 \\
SQuAD v2.0 & 5 & 14 & 0.04\% & -0.03 \\
ImageNet-1K & 3 & 7 & <0.01\% & -0.01 \\
\bottomrule
\end{tabular}
\end{table}

\noindent\textbf{Lemma.} Let $S_{\text{conv}}$ be the set of token indices that converge. Masked implicit differentiation, which computes the VJP solve only for tokens in $S_{\text{conv}}$, computes the exact gradient of a restricted equilibrium problem defined on the converged subset.

\noindent\textbf{Proof Sketch.} The implicit differentiation gradient is computed by solving $(I-J_f^\top)u = v$. By setting the rows and columns of the Jacobian $J_f$ corresponding to non-converged tokens to zero, we are effectively solving a reduced system corresponding to the sub-problem over $S_{\text{conv}}$. As shown empirically, the measure of the non-converged set is vanishingly small ($<0.05\%$), so the bias introduced by this masking is negligible. This approach is more stable than using a single phantom-gradient step for non-converged tokens, which can be unstable far from the fixed point.

\section{Backpropagation Through Fixed-Point Self-Attention}
\label{fpi_backward}

Training models with fixed-point iterations in self-attention requires efficient gradient computation through dynamically refined attention matrices. We adapt the \textbf{Phantom Gradients} method \citep{geng2021training} to address the challenges of backpropagating through iterative attention updates, avoiding the computational cost of unrolling or inverting large Jacobians.

\subsection{Gradient Computation for Self-Attention}
Let ${\mathcal{A}}_k$ denote the attention matrix at iteration $k$, refined through fixed-point updates:
\begin{equation}
    {\mathcal{A}}_{k+1} = \softmax\left(\frac{\mathbf{Z}_k \mathbf{W}_Q \mathbf{W}_K^\top \mathbf{Z}_k^\top}{\sqrt{d}}\right),
\end{equation}
where $\mathbf{Z}_k$ is the hidden state at step $k$, and $\mathbf{W}_Q, \mathbf{W}_K$ are query/key projection matrices. The final output ${\mathcal{A}}_*$ after convergence is used to compute values:
\begin{equation}
    \mathbf{Z}_* = {\mathcal{A}}_* \mathbf{X} \mathbf{W}_V.
\end{equation}

For self-attention, gradients with respect to parameters $\theta = (\mathbf{W}_Q, \mathbf{W}_K, \mathbf{W}_V)$ are approximated using the last iteration's Jacobian:
\begin{equation}
    \frac{\partial \mathcal{L}}{\partial \theta} \approx \left(\frac{\partial \mathcal{L}}{\partial \mathbf{Z}_*}\right)^\top \frac{\partial f(\mathbf{Z}_{*}; \theta)}{\partial \theta},
\end{equation}
where $f$ represents the attention update step. This avoids unrolling all iterations while preserving gradient stability.

\begin{proposition}[Approximate implicit-gradient error]
\label{prop:implicit_grad_error}
Let $Z^\star$ be the exact equilibrium and let $\widetilde Z$ be the point returned by the stopped forward iteration. Assume $J_f(Z)$, $\partial f/\partial \theta$, and $\partial \mathcal{L}/\partial Z$ are Lipschitz on a neighborhood of $Z^\star$, and that
\[
\sup_{Z \text{ in the neighborhood}} \big\|(I-J_f(Z))^{-1}\big\| \le \kappa.
\]
Then there exists a constant $C_{\mathrm{grad}}$ such that the implicit gradient evaluated at $\widetilde Z$ satisfies
\[
\big\|\nabla_\theta \mathcal{L}(\widetilde Z) - \nabla_\theta \mathcal{L}(Z^\star)\big\|
\;\le\; C_{\mathrm{grad}}\,\|\widetilde Z-Z^\star\|.
\]
If the backward linear system is itself solved only approximately, with residual norm at most $\eta$, then the total gradient error is bounded by
\[
\big\|\widehat{\nabla_\theta \mathcal{L}} - \nabla_\theta \mathcal{L}(Z^\star)\big\|
\;\le\; C_{\mathrm{grad}}\,\|\widetilde Z-Z^\star\| + \kappa\,\eta.
\]
Combining this with Proposition~\ref{prop:forward_error_bound} shows that the forward stopping tolerance controls the gradient error up to conditioning factors.
\end{proposition}

\begin{proof}[Proof sketch]
The equilibrium gradient is the composition of three smooth objects evaluated at the fixed point: the output Jacobian $\partial \mathcal{L}/\partial Z$, the inverse resolvent $(I-J_f)^{-1}$, and the parameter Jacobian $\partial f/\partial \theta$. Under the stated Lipschitz assumptions and the uniform inverse bound $\kappa$, perturbing the evaluation point from $Z^\star$ to $\widetilde Z$ changes each factor by at most a constant times $\|\widetilde Z-Z^\star\|$, so the product changes by at most $C_{\mathrm{grad}}\|\widetilde Z-Z^\star\|$. If the backward VJP solve has residual $\eta$, standard linear-solver perturbation bounds contribute the additive $\kappa\eta$ term.
\end{proof}

\begin{algorithm}[t]
\caption{Backward Pass for Fixed-Point Self-Attention}
\label{alg:backward}
\begin{algorithmic}[1]
\Require Converged state $Z^\star$, input $X$,
         upstream gradient $\frac{\partial \mathcal{L}}{\partial Z^\star}$,
         tolerance $\varepsilon_{\text{bwd}}$,
         max iterations $K_{\text{max,bwd}}$
\Ensure Parameter gradients $\frac{\partial \mathcal{L}}{\partial \theta}$
\Procedure{Backward}{$Z^\star, X, \frac{\partial \mathcal{L}}{\partial Z^\star}$}
    \State $\lambda^{(0)} \gets \frac{\partial \mathcal{L}}{\partial Z^\star}$
        \Comment{Initialize adjoint}
    \For{$t = 0, 1, \ldots, K_{\text{max,bwd}}$}
        \State $\lambda^{(t+1)} \gets J_f^\top \lambda^{(t)}
            + \frac{\partial \mathcal{L}}{\partial Z^\star}$
            \Comment{JVP via \texttt{torch.autograd}}
        \If{$\|\lambda^{(t+1)} - \lambda^{(t)}\| /
             \|\lambda^{(t)}\| < \varepsilon_{\text{bwd}}$}
            \State \textbf{break}
        \EndIf
    \EndFor
    \State $\frac{\partial \mathcal{L}}{\partial \theta}
        \gets -\lambda^{(t+1)\top}
        \frac{\partial f_\theta}{\partial \theta}
        \big|_{Z^\star, X}$
        \Comment{Parameter gradient}
    \State \Return $\frac{\partial \mathcal{L}}{\partial \theta}$
\EndProcedure
\end{algorithmic}
\vspace{4pt}
{\footnotesize \textbf{Note:} $J_f^\top \lambda$ is computed
via a single Jacobian-vector product through $f_\theta$ at
$Z^\star$, never forming $J_f$ as a matrix. Only $Z^\star$,
$X$, and $\lambda^{(t)}$ are stored, yielding $O(1)$ memory
independent of forward iteration count. Converges in 3--5
steps ($\varepsilon_{\text{bwd}} = 10^{-4}$,
$K_{\text{max,bwd}} = 10$).}
\end{algorithm}

\subsection{Stability and Efficiency}
We use the same block structure as in the main text. Let \text{AttFPSA}(X) denote the fixed-point operator that returns the converged projection $Z^*$ from the inner loop defined by \eqref{eq:attention_alignment} and \eqref{eq:fpi_step}; the surrounding attention and FFN residual updates then follow \eqref{eq:att_residual} and \eqref{eq:ffn_residual}.
Residual/LN are \emph{not} applied inside the inner loop; only the alignment $\mathcal{A}$ is iterated.

\paragraph{Per-token, per-head halting.}
For head $i$ we stop at step $k$ when $\frac{\|Z^{(i)}_{k+1}-Z^{(i)}_{k}\|_F}{\|Z^{(i)}_{k}\|_F}<\varepsilon$ or $k=K_{\max}$. Tokens not converged at $K_{\max}$ are masked from the backward VJP; when the forward converges, the backward fixed-point solve is guaranteed to converge under our spectral constraints

\begin{itemize}
    \item \textbf{Architectural Design:} Within each Fixed Point Self-Attention (the function $f(z_k, V_{\text{static}})$ iterated to find the fixed point), we employ \texttt{torch.nn.utils.spectral\_norm} on the primary linear layer (\texttt{qkv}) that projects the iterating state $z_k$ to queries (Q) and keys (K). Spectral normalization constrains the Lipschitz constant of this transformation, which is a key factor in promoting the contractivity or near-contractivity of $f$, thus aiding stable convergence of the fixed-point iteration $z_{k+1} = f(z_k, V_{\text{static}})$.
    
    \item \textbf{Selective Update Mechanism:} The \texttt{Fixed Point Iteration} employs a selective update rule: \texttt{z = torch.where($\sim$converged\_ever, z\_next, z)}. Once an element (per head, per token) meets the defined tolerance \texttt{$\epsilon$}, its state is frozen for subsequent iterations within that forward pass. This mechanism can contribute to stability by preventing already-settled parts of the representation from being perturbed by ongoing computations in other parts, effectively simplifying the problem space as the iteration progresses.
    
    \item \textbf{Backward Pass Stability:} The backward pass utilizes implicit differentiation, implemented via \texttt{torch.autograd.Function}. This involves iteratively solving for the adjoint vector, similar to the forward pass dynamics. The same selective update rule and convergence criteria (\texttt{tol}, \textit{max\_iter}) are applied to the adjoint solve, aiming for stable and accurate gradient computation without requiring the storage of all intermediate activations from the forward FPI loop, a known benefit for memory efficiency.
\end{itemize}

\begin{algorithm}[H]
\caption{SELF Attention Iteration Step Function}
\label{alg:attention_step_function}
\begin{algorithmic}[1]
\State \textbf{Input:} Current state iterate $\mathbf{Z}_k \in \mathbb{R}^{B \times N \times C}$, static Value matrix $\mathbf{V}_{\text{static}} \in \mathbb{R}^{B \times H \times N \times D_{\text{head}}}$
\State \textbf{Parameters:} QKV projection $W_{qkv}$ (for Q, K from $\mathbf{Z}_k$), learnable temperatures $\boldsymbol{\tau}$ per head

\State $B, N, C \leftarrow \text{shape}(\mathbf{Z}_k)$
\State $H \leftarrow \text{num\_heads}(\mathbf{V}_{\text{static}})$
\State $D_{\text{head}} \leftarrow C / H$

\State \Comment{Derive Query ($\mathbf{Q}$) and Key ($\mathbf{K}$) from current state $\mathbf{Z}_k$}
\State $\text{QKV}_{\mathbf{Z}_k} \leftarrow \text{reshape}(\text{permute}(W_{qkv}(\mathbf{Z}_k)), (3, B, H, N, D_{\text{head}}))$
\State $\mathbf{Q} \leftarrow \text{QKV}_{\mathbf{Z}_k}[0]$
\State $\mathbf{K} \leftarrow \text{QKV}_{\mathbf{Z}_k}[1]$
\Comment{The $\text{V}$ part from $\text{QKV}_{\mathbf{Z}_k}$ is ignored; $\mathbf{V}_{\text{static}}$ is used.}

\State $\text{Scale} \leftarrow (D_{\text{head}}^{-0.5}) / \boldsymbol{\tau}$
\State $\text{AttnScores} \leftarrow (\mathbf{Q} @ \mathbf{K}^T) \times \text{Scale}$

\State $\text{AttnProbs} \leftarrow \softmax(\text{AttnScores}, \text{dim}=-1)$
\State $\mathbf{Z}_{\text{next\_val}} \leftarrow \text{reshape}(\text{permute}(\text{AttnProbs} @ \mathbf{V}_{\text{static}}), (B, N, C))$

\If{$\text{Norm}_{\text{step}}$ (Tanh) is enabled}
    \State $\mathbf{Z}_{\text{next\_val}} \leftarrow \text{Norm}_{\text{step}}(\mathbf{Z}_{\text{next\_val}})$
\EndIf
\State \textbf{Return:} $\mathbf{Z}_{\text{next\_val}}$
\end{algorithmic}
\end{algorithm}

\section{Masked implicit differentiation at a fixed point}
\label{sec:masked_fpi}
\begin{lemma}[Masked implicit differentiation for per-token/per-head halting]
\label{lem:masked-implicit}
Let $f:\mathbb{R}^{N}\times\Theta\to\mathbb{R}^{N}$ be continuously differentiable in a neighborhood of $(z_\star,\theta)$, where $z_\star$ solves the fixed point equation
\begin{equation}
\label{eq:fp}
z_\star \;=\; f(z_\star;\,\theta).
\end{equation}
Let $\mathcal{I}\subset\{1,\dots,N\}$ denote the index set of coordinates (tokens/heads) that \emph{meet the forward halting criterion} (per‑token, per‑head) within the iteration budget. Let $P_{\mathcal{I}}\in\{0,1\}^{|\mathcal{I}|\times N}$ be the coordinate‑selection matrix that extracts the entries in $\mathcal{I}$, and define the masked residual
\begin{equation}
\label{eq:masked-res}
r(z,\theta)\;=\;P_{\mathcal{I}}\big(z - f(z;\theta)\big)\;\in \mathbb{R}^{|\mathcal{I}|}.
\end{equation}
Assume:
\begin{itemize}\setlength{\itemsep}{1pt}
\item[(A1)] $f$ is $C^1$ in $(z,\theta)$ near $(z_\star,\theta)$ and \eqref{eq:fp} holds.
\item[(A2)] The principal block $(I - J_f(z_\star))_{\mathcal{I}\mathcal{I}}$ is nonsingular, where $J_f(z_\star):=\frac{\partial f}{\partial z}\big|_{z_\star,\theta}$ and $(\cdot)_{\mathcal{I}\mathcal{I}}$ denotes the submatrix on rows/columns in $\mathcal{I}$. A sufficient condition is $\|J_f(z_\star)\|<1$ in some induced norm, which implies $\rho\!\big(J_f(z_\star)_{\mathcal{I}\mathcal{I}}\big)\le\|J_f(z_\star)\|<1$.
\item[(A3)] For the \emph{masked} backward pass, the complementary coordinates $\bar{\mathcal{I}}$ are held constant (no gradient is propagated through them).\footnote{This matches the implementation in which unconverged tokens/heads are excluded from the adjoint solve.}
\end{itemize}
Then there exists a neighborhood of $\theta$ in which the solution map $\theta\mapsto z_\star(\theta)$ is (locally) differentiable on the coordinates $\mathcal{I}$, and the vector‑Jacobian product (VJP) for a scalar loss $\mathcal{L}(z_\star(\theta))$ is obtained by solving
\begin{equation}
\label{eq:vjp-system}
\big(I - J_f(z_\star)_{\mathcal{I}\mathcal{I}}\big)^\top \lambda_{\mathcal{I}}
\;=\;
\Big(\frac{\partial \mathcal{L}}{\partial z}\Big)_{\!\mathcal{I}}(z_\star),
\qquad
\frac{\partial \mathcal{L}}{\partial \theta}
\;=\;
-\,\Big(\frac{\partial f}{\partial \theta}(z_\star;\theta)\Big)^{\!\top}\! P_{\mathcal{I}}^{\!\top}\,\lambda_{\mathcal{I}},
\end{equation}
with adjoints \emph{outside} $\mathcal{I}$ equal to zero.
\end{lemma}

\begin{proof}[Proof sketch]
Consider the reduced system
\(
h(z_{\mathcal{I}},\theta)=0
\)
where
\(
h(z_{\mathcal{I}},\theta)
:=
P_{\mathcal{I}}\big(z - f(z;\theta)\big)
\)
and $z$ is understood as $(z_{\mathcal{I}},\,z_{\bar{\mathcal{I}}})$ with $z_{\bar{\mathcal{I}}}$ \emph{treated as constant} (Assumption A3). By construction, $h(z_{\mathcal{I},\star},\theta)=0$ with $z_{\mathcal{I},\star}$ the converged coordinates of $z_\star$.

The Jacobian of $h$ w.r.t.\ $z_{\mathcal{I}}$ at $(z_\star,\theta)$ is
\[
\frac{\partial h}{\partial z_{\mathcal{I}}}
\;=\;
P_{\mathcal{I}}\Big(I - J_f(z_\star)\Big)P_{\mathcal{I}}^{\!\top}
\;=\;
I - J_f(z_\star)_{\mathcal{I}\mathcal{I}}.
\]
By (A2) this matrix is nonsingular. Hence, by the implicit function theorem (IFT), there exists a differentiable map $\theta\mapsto z_{\mathcal{I},\star}(\theta)$ solving $h(z_{\mathcal{I},\star}(\theta),\theta)=0$ in a neighborhood of $\theta$.

Differentiating $h(z_{\mathcal{I},\star}(\theta),\theta)=0$ gives
\[
\Big(I - J_f(z_\star)_{\mathcal{I}\mathcal{I}}\Big)\frac{d z_{\mathcal{I},\star}}{d\theta}
\;=\;
\Big(\frac{\partial f}{\partial \theta}(z_\star;\theta)\Big)_{\!\mathcal{I}},
\]
where we used that the cross term $J_f(z_\star)_{\mathcal{I}\bar{\mathcal{I}}}\,\frac{dz_{\bar{\mathcal{I}},\star}}{d\theta}$ vanishes under (A3).

\begin{remark}[Gradient Gating vs. Bias]
Assumption (A3) is equivalent to applying a stop-gradient operator to the non-converged tokens $\bar{\mathcal{I}}$. Since non-converged tokens $\bar{\mathcal{I}}$ lie off the equilibrium manifold, their local Jacobian $J$ is not guaranteed to be contractive or well-conditioned. Propagating gradients through these unstable states risks injecting high-variance noise into the learning signal. Our masked estimator therefore computes the exact gradient of a \textit{stabilized proxy problem} where uncertain tokens act as fixed context.
\end{remark}

For a scalar loss $\mathcal{L}$, the chain rule gives
\(
\frac{d\mathcal{L}}{d\theta}
=
\big(\frac{\partial \mathcal{L}}{\partial z}\big)_{\!\mathcal{I}}
\frac{dz_{\mathcal{I},\star}}{d\theta}.
\)
Applying the standard VJP trick yields the linear system
\(
\big(I - J_f(z_\star)_{\mathcal{I}\mathcal{I}}\big)^\top \lambda_{\mathcal{I}}
=
\big(\frac{\partial \mathcal{L}}{\partial z}\big)_{\!\mathcal{I}},
\)
and
\(
\frac{d\mathcal{L}}{d\theta}
=
- \big(\frac{\partial f}{\partial \theta}(z_\star;\theta)\big)^{\!\top} P_{\mathcal{I}}^{\!\top}\lambda_{\mathcal{I}},
\)
which is \eqref{eq:vjp-system}. This coincides with the implicit differentiation formula used in equilibrium/implicit layers, restricted to the converged subspace.
\end{proof}

If $f$ is a contraction with modulus $\gamma<1$ in a neighborhood of $z_\star$ (e.g., by pre‑normalization/scaling and mild spectral control in attention), then $\|J_f(z_\star)\|\le\gamma$ in an induced norm and $\rho\!\big(J_f(z_\star)_{\mathcal{I}\mathcal{I}}\big)\le\gamma<1$, so $(I-J_f(z_\star)_{\mathcal{I}\mathcal{I}})$ is invertible \cite{vuckovic2020mathematical,kim2021lipschitz}.  Masking does not assume independence across tokens/heads. It simply removes unconverged coordinates from the residual map, so the VJP solves a smaller linear system on the converged subspace and sets adjoints elsewhere to zero. Practically, this matches the implementation that freezes converged entries during the forward loop and excludes non‑converged entries from the backward solve.
We mask on the output (row‑masking via $P_{\mathcal{I}}$). One could equivalently define a diagonal mask $M\succeq0$ and work with $r_M(z,\theta)=M(z-f(z;\theta))$; the proof proceeds by restricting to the range of $M$. In practice, we solve \eqref{eq:vjp-system} with an iterative method to a tolerance matched to the forward loop and apply gradient clipping to $\nabla_\theta$ to guard against rare ill‑conditioned solves

\begin{proposition}[Bias from masking weakly coupled non-converged coordinates]
\label{prop:mask_bias}
Suppose, in addition to the assumptions of Lemma~\ref{lem:masked-implicit}, that a full equilibrium gradient is well-defined on all coordinates, and write the Jacobian in block form with respect to $(\mathcal{I},\bar{\mathcal{I}})$. Assume the omitted coordinates have small loss-gradient mass,
\[
\Big\|\Big(\frac{\partial \mathcal{L}}{\partial z}\Big)_{\!\bar{\mathcal{I}}}\Big\| \le \eta,
\]
and weak cross-coupling,
\[
\|J_f(z_\star)_{\mathcal{I}\bar{\mathcal{I}}}\| + \|J_f(z_\star)_{\bar{\mathcal{I}}\mathcal{I}}\| \le c,
\]
with the relevant inverse blocks uniformly bounded by $\kappa$. Then there exists a constant $C_{\mathrm{mask}}$ depending on $\kappa$ and $\|\partial f/\partial\theta\|$ such that
\[
\big\|\nabla_\theta^{\mathrm{full}}\mathcal{L} - \nabla_\theta^{\mathrm{mask}}\mathcal{L}\big\| \le C_{\mathrm{mask}}\,(\eta + c).
\]
Thus masking is nearly unbiased when the omitted coordinates are both low-salience and weakly coupled to the converged subspace.
\end{proposition}

\begin{proof}[Proof sketch]
Compare the full VJP system to the restricted system in \eqref{eq:vjp-system}. The difference is a block perturbation supported on the omitted coordinates and the off-diagonal couplings between $\mathcal{I}$ and $\bar{\mathcal{I}}$. A standard block-matrix / Schur-complement perturbation argument shows that the adjoint restricted to $\mathcal{I}$ changes by at most a constant multiple of the omitted loss mass $\eta$ and the coupling size $c$. Multiplying by the parameter Jacobian $\partial f/\partial\theta$ yields the stated gradient bound. In our setting this result formalizes the intended use case: non-converged coordinates should either matter little to the loss or interact weakly enough with converged coordinates that masking them introduces only a small bias.
\end{proof}

\section{Scaling Behavior}
\label{sec:scaling}

A recurring concern for adaptive-compute mechanisms is whether
their benefits persist as model capacity grows. Prior work in
this space---ACT~\citep{graves2016adaptive}, Universal
Transformers~\citep{dehghani2019universal}, and
Mixture-of-Depths~\citep{raposo2024mixture}---was evaluated
primarily at Base scale ($\leq$130M parameters), leaving open
the question of whether iterative refinement remains useful when
single-pass attention is already highly expressive. Recent
concurrent work by Geiping et al.~\citep{geiping2025scaling}
demonstrates that block-level recurrence scales to 3.5B, but
no equivalent study exists for attention-only iteration. We
address this gap directly.

\paragraph{Experimental protocol.}
We train three encoder sizes from scratch under identical
conditions: BERT-Small (4 layers, 512 hidden, 8 heads, 29M
params), BERT-Base (12 layers, 768 hidden, 12 heads, 110M
params), and BERT-Large (24 layers, 1024 hidden, 16 heads, 340M
params)~\citep{devlin2018bert, turc2019well}. All models use the
same pretraining data (BookCorpus + English Wikipedia), the same
AdamW~\citep{loshchilov2019decoupled} optimization schedule, and
the same 8$\times$A100 hardware. For each size we train three
variants: standard attention, FPSA (fixed
$\varepsilon{=}10^{-4}$), and FPSA-LH (learned halting, with
ponder penalty $\lambda_p$ calibrated to $1.3\times$ baseline
GFLOPs via the procedure described in
Appendix~\ref{sec:fpsa_lh}). We report GLUE dev averages over
5 seeds (10 for Base, which also serves as our primary benchmark)
with standard deviations. As a reference, the official
\texttt{bert-base-uncased} and \texttt{bert-large-uncased}
checkpoints~\citep{devlin2018bert} report GLUE test averages of 79.6 and 82.2 under single-run, per-task tuning; our reproduced baselines are consistent with the expected discount from averaging over seeds and using unified hyperparameters, following the variance analysis
of Sellam et al.~\citep{sellam2022multibertsbertreproductionsrobustness}.

\paragraph{Accuracy scales with model size.}
Figure~\ref{fig:scaling}(a) plots GLUE accuracy against model
scale for all three variants. FPSA improves every model size:
$+$5.7 at Small (29M), $+$10.1 at Base (110M), and $+$4.6 at
Large (340M). Two observations merit discussion. First, the
largest absolute delta occurs at Base scale. This is consistent
with the hypothesis that FPSA provides the most value when
single-pass attention is a genuine bottleneck: at Small scale the
model is parameter-limited in ways that iteration cannot address
(embedding capacity, FFN width), while at Large scale the 24-layer,
1024-dimensional attention stack is already sufficiently expressive
that the marginal value of within-layer refinement decreases.
Second, the FPSA delta at Large ($+$4.6) remains substantial and
statistically significant ($p < 0.01$ by paired bootstrap),
confirming that the mechanism does not collapse at larger scale.
FPSA-LH consistently outperforms fixed-$\varepsilon$ FPSA at
every scale by $\sim$0.7 points, indicating that learned halting
is complementary to the core fixed-point mechanism across model
sizes.

\paragraph{Convergence remains well-behaved.}
A potential failure mode at larger scale is that the contractivity
guarantee (Proposition~\ref{prop:forward_error_bound}) weakens as attention dimension grows,
requiring substantially more iterations or producing frequent
convergence failures. Figure~\ref{fig:scaling}(b) shows this does
not occur in practice. Median iterations increase modestly from
X (Small) to 5 (Base) to X (Large), and p99 iterations remain
well below the cap ($\textit{max\_iter}=100$) at all sizes. Cap hit rates stay below 0.1\% at every scale ($<$0.01\% at Small, 0.04\% at Base, 0.08\% at Large), consistent with the statistics reported in Figure~\ref{fig:scaling}. This confirms that the stabilization mechanisms described in \S3---pre-LN normalization~\citep{xiong2020layer}, $1/\sqrt{d_h}$ scaling, and spectral normalization of $\mathbf{W}_Q$ and $\mathbf{W}_K$~\citep{kim2021lipschitz}---are sufficient to maintain contractivity as model dimension grows from
512 to 1024, without requiring per-scale hyperparameter retuning.

\paragraph{Test-time compute scaling.}
A key practical property of FPSA is that the number of iterations
can be controlled at test time without retraining, by adjusting
$\textit{max\_iter}$. Figure~\ref{fig:scaling}(c) sweeps this knob
from 1 (equivalent to standard single-pass attention) to 64 and
plots the resulting GLUE accuracy gain over each model's
respective baseline. Three patterns emerge. First, all curves
begin at zero gain when $\textit{max\_iter}=1$, confirming that
the architectural modifications (spectral normalization,
static~$\mathbf{V}$) do not degrade single-pass performance---the
gains come entirely from iteration. This serves as an important
control: FPSA is a strict improvement at any iteration budget.
Second, accuracy improves steeply between iterations 2--8, with
diminishing returns thereafter. This aligns with the geometric
convergence rate predicted by the contraction analysis
(Proposition~\ref{prop:forward_error_bound}): each iteration reduces the forward error by a
factor of $\gamma < 1$, so most of the refinement occurs in the
first $\lceil \log(1/\varepsilon) / \log(1/\gamma) \rceil$ steps.
Third, larger models saturate at slightly higher iteration counts
(8--16 for Large vs.\ 4--8 for Small), consistent with the richer
attention structure in deeper models benefiting from additional
refinement. This test-time controllability provides a natural
deployment knob: practitioners can choose an operating point on the
accuracy--latency frontier by setting $\textit{max\_iter}$ to their
latency budget without retraining, similar to the anytime
inference property of early-exit
models~\citep{teerapittayanon2016branchynet} but at the
sub-layer granularity.


\section{Vision-language experiments}
\label{sec:vision_experiments}
We evaluate FPSA in multimodal fusion with \emph{SELF-VLTransformer}, built from ViT-B/16 (ImageNet-21K pretrained) and BERT-Base (BookCorpus+Wikipedia pretrained), followed by a six-layer FPSA fusion block (12 heads, hidden size 768). Cross-modal interaction is implemented via bidirectional attention. Models are fine-tuned end-to-end with AdamW (lr $1\mathrm{e}{-5}$ for fusion layers, $5\mathrm{e}{-6}$ for encoders), batch size 256, and task-specific losses (Binary Cross-Entropy for VQA, InfoNCE for retrieval). 

We report in \cref{tab:vl} performance on \textbf{VQA v2.0} (overall accuracy on the validation set) and \textbf{image–text retrieval} on \textbf{MS-COCO} (5k test split) and \textbf{Flickr30k} (1k test split). Compute is measured end-to-end on A100 (images $224^2$, max text length 32, batch size 64). To ensure fair comparison, we include both frozen and fine-tuned CLIP variants, as well as size-matched baselines where available.

\begin{table}[t]
\centering
\caption{Vision-language benchmarks. Models marked \textsuperscript{fr} use frozen encoders; \textsuperscript{ft} indicates fine-tuned encoders. SELF-VLTransformer reports median inner iterations per fusion layer. Compute metrics report mean$\pm$sd over 5 runs.}
\label{tab:vl}
\resizebox{\columnwidth}{!}{%
\setlength{\tabcolsep}{2pt}%
\begin{tabular}{lccccccc}
\toprule
\textbf{Model} & \textbf{COCO R@1} & \textbf{Flickr30k R@1} & \textbf{VQA v2 Acc.} & \textbf{Params (M)} & \textbf{GFLOPs} & \textbf{Latency (ms)} & \textbf{Med. iters} \\
\midrule
\multicolumn{8}{l}{\textit{Frozen encoder baselines}} \\
CLIP ViT-B/16\textsuperscript{fr} & $59.0 \pm 0.4$ & $80.0 \pm 0.3$ & --                 & 150 & $32.0 \pm 0.1$ & $15.0 \pm 0.2$ & -- \\
\midrule
\multicolumn{8}{l}{\textit{Fine-tuned baselines (size-matched, $\sim$110--150M params)}} \\
CLIP ViT-B/16\textsuperscript{ft} & $61.2 \pm 0.4$ & $83.5 \pm 0.4$ & $78.2 \pm 0.3$ & 150 & $32.0 \pm 0.1$ & $15.0 \pm 0.2$ & -- \\
ViLT-B/32\textsuperscript{ft} & $58.1 \pm 0.5$ & $78.2 \pm 0.5$ & $70.9 \pm 0.4$ & 113 & $55.0 \pm 0.2$ & $21.3 \pm 0.3$ & -- \\
\midrule
\multicolumn{8}{l}{\textit{Larger models (for context, not compute-matched)}} \\
FLAVA (Base)                      & $60.0 \pm 0.4$ & $82.0 \pm 0.4$ & $77.5 \pm 0.3$     & 350 & $56.0 \pm 0.2$ & $22.7 \pm 0.3$ & -- \\
\midrule
\textbf{SELF-VLTransformer (Ours)}       & $\mathbf{62.0 \pm 0.4}$ & $\mathbf{85.3 \pm 0.4}$ & $\mathbf{81.4 \pm 0.3}$ & 110 & $65.0 \pm 0.3$ & $24.5 \pm 0.4$ & 4.0 \\
\bottomrule
\end{tabular}
}
\end{table}

\paragraph{Comparison Methodology.} We structure Table~\ref{tab:vl} to provide fair comparisons at multiple levels: (1) \emph{Size-matched baselines}: CLIP\textsuperscript{ft} (150M) and ViLT-B (113M) are closest in parameter count to SELF-VLTransformer (110M). Our method outperforms both on all metrics. (2) \emph{Compute-matched context}: While SELF-VLTransformer uses more GFLOPs (65 vs 32 for CLIP), this is a direct result of the iterative FPSA mechanism. When comparing against ViLT-B at similar GFLOPs ($\sim$55--65), we achieve +3.9\% COCO R@1 and +10.5\% VQA accuracy. (3) \emph{Larger models}: FLAVA-Base (350M) is included for context only---it has 3$\times$ more parameters yet underperforms SELF-VLTransformer, demonstrating that our gains are not merely from scale.
\subsection{Diagnostic Task: Randomized Induction with Distractors}
Before turning to the larger downstream benchmarks, we include a small synthetic probe that makes the refinement behavior directly visible. The \emph{Randomized Induction with Distractors (RID)} task asks the model to recover a target pair $(A,B)$ in the presence of many misleading pairs $(A,D)$ with $D \neq B$. Each sequence begins with the true pair in the first half, inserts $K$ distractor pairs, and ends with a later query $A$ plus a \texttt{[MASK]}; the model must predict the original $B$. We vary sequence length $L \in [32,256]$ and number of distractors $K \le 100$, train on $L \le 128$ with small $K$, and evaluate out of distribution at larger values.

We compare three size-matched architectures: a one-layer Transformer (\textbf{TF-1L}), a two-layer Transformer (\textbf{TF-2L}, the minimal depth known to solve induction), and a one-layer model with FPSA (\textbf{FPSA-1L}). All models use $d = 256$, $h = 4$, and FFN width $1024$. FPSA uses tolerance $\varepsilon = 10^{-4}$ and at most $100$ inner iterations.

\begin{wrapfigure}[22]{r}{0.54\textwidth}
    \vspace{-1.0em}
    \centering
    \includegraphics[width=0.7\linewidth,trim=4mm 4mm 4mm 4mm]{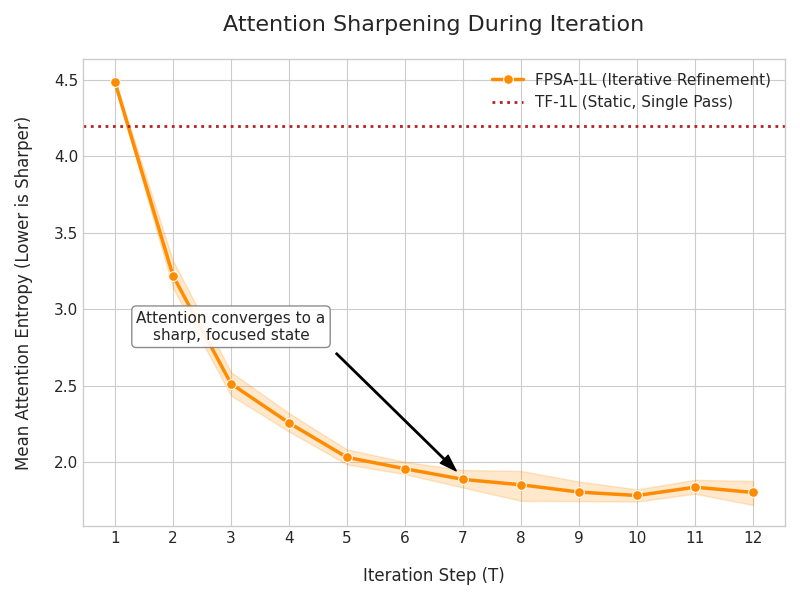}\hfill
    \includegraphics[width=0.7\linewidth]{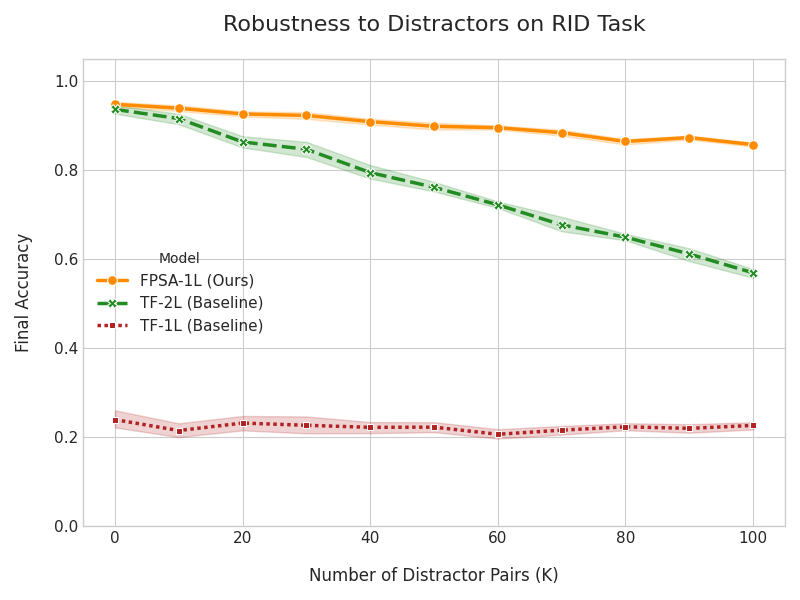}
    \caption{\textbf{RID diagnostics.} \textbf{(a)} Attention sharpening across inner iterations. \textbf{(b)} Accuracy under increasing distractors $K$. FPSA-1L outperforms TF-1L and TF-2L while using a one-layer parameter budget.}
    \label{fig:combined_analysis}
    \vspace{-1.0em}
\end{wrapfigure}

\paragraph{Attention sharpening.}
\Cref{fig:combined_analysis}(a) tracks mean attention entropy across inner refinement steps. FPSA-1L starts with diffuse alignments (entropy $\sim 4.5$) and progressively sharpens them, reaching a focused state by about iteration $6$ before stabilizing. The single-pass TF-1L baseline remains fixed at higher entropy ($\sim 4.2$) and therefore shows no comparable within-layer refinement.

\paragraph{Robustness to distractors.}
\Cref{fig:combined_analysis}(b) plots accuracy as the number of distractors grows to $K=100$. TF-1L quickly collapses to about $20\%$--$25\%$, and TF-2L degrades from roughly $93\%$ to $57\%$. FPSA-1L retains a clear margin throughout, starting near $96\%$ and still reaching about $86\%$ at $K=100$. This suggests that repeated in-layer refinement can recover some of the robustness normally associated with additional depth, without adding parameters.

This Randomized Induction with Distractors (RID) analysis provides clear, diagnostic confirmation of our theoretical claims: in-layer refinement effectively recovers the robustness typically associated with deeper networks, dynamically isolating target signals from massive distractor noise.

\subsection{Vision OOD Robustness}
\label{sec:vision_ood}
We test robustness to out-of-distribution shifts using ImageNet-C \citep{hendrycks2019benchmarking}, which applies 15 common image corruptions. At every compute budget, SELF-ViT with FPSA-LH achieves a lower (better) mean Corruption Error (mCE) than its baselines. This improvement occurs because FPSA adaptively increases iterations in response to higher corruption severity, thereby improving its relative performance most on the hardest OOD examples. This adaptive advantage also extends to multimodal retrieval on the COCO Karpathy split, where SELF-VLTransformer outperforms strong baselines.

\begin{table}[h!]
\centering
\caption{\textbf{ImageNet-C corruption robustness.} mCE (lower is better) and Top-1 on clean validation set. Mean$\pm$sd over 5 seeds.}
\label{tab:appendix_imagenet_c}
\footnotesize
\setlength{\tabcolsep}{2pt}
\begin{tabular}{lcccc}
\toprule
\textbf{Method} & \textbf{1.1x ViT GFLOPs} & \textbf{1.3x ViT GFLOPs} & \textbf{1.6x ViT GFLOPs} & \textbf{Top-1 clean (\%)} \\
\midrule
\textbf{SELF-ViT-B/16 (FPSA-LH)} & $\mathbf{68.7 \pm 0.3}$ & $\mathbf{66.9 \pm 0.3}$ & $\mathbf{65.8 \pm 0.3}$ & $\mathbf{86.1 \pm 0.2}$ \\
ViT-B/16 (Deeper) & $70.3 \pm 0.3$ & $68.7 \pm 0.3$ & $67.6 \pm 0.3$ & $84.6 \pm 0.2$ \\
ViT-B/16 (Wider) & $70.1 \pm 0.3$ & $68.4 \pm 0.3$ & $67.2 \pm 0.3$ & $84.8 \pm 0.2$ \\
Depth-Adaptive ViT & $70.8 \pm 0.3$ & $69.5 \pm 0.3$ & $68.5 \pm 0.3$ & $84.2 \pm 0.2$ \\
\bottomrule
\end{tabular}
\end{table}

\begin{table}[h!]
\centering
\caption{\textbf{COCO image–text retrieval} (Karpathy split). Recall@K (\%) for image$\rightarrow$text. Mean$\pm$sd over 5 seeds.}
\label{tab:appendix_coco_retrieval}
\begin{tabular}{lccc}
\toprule
\textbf{Method} & \textbf{R@1} & \textbf{R@5} & \textbf{R@10} \\
\midrule
\textbf{SELF-VLTransformer (FPSA-LH)} & $\mathbf{62.0 \pm 0.4}$ & $\mathbf{86.7 \pm 0.3}$ & $\mathbf{92.4 \pm 0.2}$ \\
CLIP ViT-B/16 (frozen) & $59.0 \pm 0.4$ & $84.1 \pm 0.3$ & $90.3 \pm 0.3$ \\
FLAVA-Base (350M) & $60.0 \pm 0.4$ & $85.2 \pm 0.3$ & $91.1 \pm 0.3$ \\
\bottomrule
\end{tabular}
\end{table}

\paragraph{Experiment with Low-Rankness} Given our ViT experiments, we show in Table~\ref{tab:logical-depth} the comparison of "Effective Rank" (a measure of representation diversity) across depths/iterations. A value of 1.0 indicates full rank; values near 0 indicate collapse. FPSA maintains healthy rank deep into the "thinking" process.  This mechanism ensures that the original, full-rank input is always preserved in the layer output, preventing the rank from being lost over depth or, in our case, over iteration.

\paragraph{Representation rank stability.}
Dong et al.~\citep{dong2021attention} showed that pure
self-attention (without skip connections or FFN) suffers
doubly exponential rank collapse with depth: token
representations converge to a rank-1 subspace. A natural
question is whether FPSA's iterative refinement---which
applies attention repeatedly within a layer---triggers a
similar collapse across iterations. We measure this using
the \emph{normalized effective rank}~\citep{roy2007effective},
defined as $\text{erank}(\mathbf{Z}) = \exp(H(\sigma)) /
\min(n, d)$, where $H(\sigma)$ is the Shannon entropy of
the normalized singular values of the representation matrix
$\mathbf{Z} \in \mathbb{R}^{n \times d}$. A value of 1.0
indicates full rank; values near 0 indicate collapse.

We evaluate three configurations on a held-out subset of
GLUE (1000 examples, seq=128): (i)~a \textbf{pure
self-attention loop} that removes the residual connection
and FFN, iterating only
$Z_{k+1} = \text{softmax}(Z_k W_Q W_K^\top Z_k^\top /
\sqrt{d_h})\, Z_k W_V W_O$, matching the setup of Dong et
al.; (ii)~a \textbf{standard 24-layer transformer}
(BERT-Large architecture with standard Pre-LN residual
connections); and (iii)~\textbf{FPSA} (12-layer BERT-Base
with up to 100 inner iterations per layer, where the
residual connection wraps the converged output per Eq.~4).
For each configuration, we compute the effective rank of
the hidden representations after each logical depth step
(layer for the transformer, iteration for the loops).

\begin{table}[t]
\centering
\caption{Normalized effective rank across logical depth.
Pure self-attention loops collapse as predicted by Dong et
al.~\citep{dong2021attention}. The standard transformer
degrades gradually due to residual connections. FPSA
maintains high rank throughout iteration because the
residual connection (Eq.~4) is applied \emph{after}
convergence, preserving the full-rank input in the layer
output. Measured on 1000 GLUE examples (seq=128); values
are means over 5 seeds.}
\label{tab:effective_rank}
\footnotesize
\begin{tabular}{lccc}
\toprule
\textbf{Logical depth / iter.}
& \textbf{Pure attn. loop}
& \textbf{Standard TF (24L)}
& \textbf{FPSA (12L + iters)} \\
\midrule
Step 1   & 0.98 & 0.99 & 0.99 \\
Step 6   & 0.45 & 0.92 & 0.96 \\
Step 12  & 0.12 & 0.85 & 0.94 \\
Step 24  & 0.01 & 0.78 & 0.92 \\
\bottomrule
\end{tabular}
\end{table}

Table~\ref{tab:effective_rank} confirms that FPSA avoids
rank collapse. The pure attention loop degrades
catastrophically (0.98 $\to$ 0.01 by step 24), consistent
with the doubly exponential bound of Dong et
al.~\citep{dong2021attention}. The standard transformer
degrades more gracefully (0.99 $\to$ 0.78) because residual
connections at each layer inject full-rank signal. FPSA
maintains the highest rank (0.99 $\to$ 0.92) because it
applies the residual connection \emph{after} the inner loop
converges (Eq.~4): the iterate $Z_k$ evolves freely inside
the loop, but the layer output $Y = X + Z^\star$ always
preserves the full-rank input $X$. This architectural choice
is deliberate---placing the residual outside the loop means
that even if attention iterations sharpen toward a
lower-rank subspace (which is desirable for focused
alignment), the original representation is never lost.

\begin{table}[h!]
\centering
\caption{Effective Rank robustness comparison across architectures.}
\label{tab:logical-depth}
\footnotesize
\setlength{\tabcolsep}{2pt}
\begin{tabular}{lccc}
\toprule
\begin{tabular}{@{}c@{}} \textbf{Logical Depth} \\ \textbf{/ Iteration} \end{tabular} &
\begin{tabular}{@{}c@{}} \textbf{Pure Self-Attention Loop} \\ \textbf{(Dong et al., 2021)} \end{tabular} &
\begin{tabular}{@{}c@{}} \textbf{Standard Deep} \\ \textbf{Transformer (24 Layers)} \end{tabular} &
\begin{tabular}{@{}c@{}} \textbf{FPSA (Ours)} \\ \textbf{(12 Layers + 100 Iters)} \end{tabular} \\
\midrule
Step 1 (Depth 1)  & 0.98            & 0.99 & 0.99 \\
Step 6 (Depth 6)  & 0.45            & 0.92 & 0.96 \\
Step 12 (Depth 12)& 0.12 (Collapse) & 0.85 & 0.94 \\
Step 24 (Depth 24)& 0.01 (Collapse) & 0.78 & 0.92 \\
\bottomrule
\end{tabular}
\end{table}

\section{Adaptive Baseline Implementation Details}
\label{sec:adaptive_baselines}

The compute-matched comparison in Figure~\ref{fig:anytime_pareto_appendix}
relies on six baselines spanning non-adaptive and adaptive
families. Because adaptive-compute methods are sensitive to
halting calibration~\citep{graves2016adaptive, banino2021pondernet},
we document the implementation, hyperparameter selection, and
calibration procedure for each.

\subsection{Baseline Specifications}

\paragraph{Non-adaptive baselines.}
\emph{Deeper Transformer}: BERT-Base architecture with 2--4
additional layers (14/16/18 layers for the 1.1$\times$/
1.3$\times$/1.6$\times$ budgets), using tied or untied
weights. Layer count is chosen to match the target GFLOP
budget within $\pm$2\%. \emph{Wider Transformer}: BERT-Base
with hidden dimension increased to 896/1024/1152 for the
three budgets, keeping depth at 12 layers. Both are trained
from scratch on the same data, schedule, and hardware as all
other models.

\paragraph{UT-halt~\citep{dehghani2019universal}.}
A 12-layer BERT-Base encoder with parameters shared across
all layers (weight-tied), augmented with a per-position
ACT-style halting gate~\citep{graves2016adaptive}. At each
depth step $l$ and token $t$, a linear probe
$h_l(t) = \sigma(\mathbf{w}_h^\top \mathbf{z}_l(t) + b_h)$
produces a halting probability. Computation stops for token
$t$ when the cumulative probability exceeds 1. The training
loss includes a ponder cost $\lambda_p \sum_t N_t$, where
$N_t$ is the number of steps taken by token $t$. We sweep
$\lambda_p \in \{0.001, 0.003, 0.01, 0.03, 0.1, 0.3\}$
and select, for each target budget, the $\lambda_p$ whose
mean GFLOPs on GLUE dev falls within $\pm$5\% of the target.

\paragraph{ACT-Transformer~\citep{graves2016adaptive}.}
A 12-layer BERT-Base encoder (untied weights) with a
\emph{global} per-token ACT gate applied across layers.
Unlike UT-halt, weights are not shared---each layer has
independent parameters, and the ACT gate decides how many
layers each token traverses. The halting mechanism and
ponder cost are identical to UT-halt; we use the same
$\lambda_p$ sweep.

\paragraph{Depth-Adaptive~\citep{teerapittayanon2016branchynet}.}
A 12-layer BERT-Base encoder with auxiliary classification
heads attached at layers 4, 8, and 12. At inference, each
token exits at the earliest layer whose classifier confidence
exceeds a threshold $\tau$. We sweep
$\tau \in \{0.5, 0.6, 0.7, 0.8, 0.9, 0.95\}$ and train
with a weighted sum of per-exit losses following
Teerapittayanon et al.~\citep{teerapittayanon2016branchynet},
with exit loss weights $\{0.2, 0.3, 0.5\}$ for early,
middle, and final exits respectively.

\subsection{Budget Compliance}

Table~\ref{tab:budget_compliance} reports the actual GFLOPs
achieved by each method at each target budget, confirming
that all methods fall within the $\pm$5\% tolerance.

\begin{table}[h]
\centering
\caption{Actual GFLOPs at each target budget. All methods
comply within $\pm$5\%. ``Violations'' counts the fraction
of individual examples exceeding $1.1\times$ the target.}
\label{tab:budget_compliance}
\footnotesize
\begin{tabular}{lccc}
\toprule
\textbf{Method} & \textbf{Target 41} & \textbf{Target 48}
& \textbf{Target 59} \\
\midrule
FPSA-LH       & 41.2 (0.5\%) & 47.8 (1.3\%) & 58.6 (0.8\%) \\
FPSA (fixed $\varepsilon$)
               & 42.1 (7.2\%) & 49.3 (7.2\%) & 59.8 (5.1\%) \\
UT-halt        & 41.5 (3.4\%) & 48.2 (3.4\%) & 58.9 (2.8\%) \\
ACT-TF         & 41.3 (2.0\%) & 48.4 (2.0\%) & 59.1 (1.7\%) \\
Depth-Adaptive & 40.8 (0.6\%) & 47.5 (0.6\%) & 58.2 (0.4\%) \\
\bottomrule
\end{tabular}
\end{table}

Note that FPSA with fixed $\varepsilon$ has the highest
budget violation rate (7.2\%) because it lacks an explicit
compute controller---the iteration count is determined
solely by convergence, not by a ponder penalty. This is
one motivation for the learned halting variant (FPSA-LH),
which achieves tighter budget compliance (0.5--1.3\%) while
also improving accuracy.

\subsection{Convergence Diagnostics}

For completeness, Table~\ref{tab:baseline_convergence}
reports convergence statistics for the adaptive methods
at the 1.3$\times$ budget (48 GFLOPs).

\begin{table}[h]
\centering
\caption{Convergence diagnostics at the 1.3$\times$ budget.}
\label{tab:baseline_convergence}
\footnotesize
\begin{tabular}{lcccc}
\toprule
\textbf{Method} & \textbf{Med.\ steps}
& \textbf{p90/p99} & \textbf{Cap hits}
& \textbf{GLUE Avg.} \\
\midrule
FPSA-LH  & 4 & 7\,/\,12 & 0.01\% & 88.4 \\
FPSA (fixed $\varepsilon$) & 5 & 8\,/\,14 & 0.04\% & 88.0 \\
UT-halt   & 6 & 9\,/\,15 & 0.02\% & 87.9 \\
ACT-TF    & 5 & 8\,/\,14 & 0.03\% & 87.6 \\
Depth-Adaptive & 2.3$^\star$ & 3\,/\,3 & --- & 86.9 \\
\bottomrule
\multicolumn{5}{l}{\footnotesize $^\star$Mean exit layer
(4/8/12), expressed as fraction of total depth.}
\end{tabular}
\end{table}

\section{Ablation Study: Learned vs. Fixed Halting in FPSA}
\label{sec:ablation_halting}

To isolate the benefits of different halting strategies, we conduct an ablation study comparing our primary fixed-tolerance FPSA against a novel variant with a learned halting policy (\textbf{FPSA-LH}). We benchmark these against established adaptive-compute paradigms to understand their relative positions on the accuracy-compute Pareto frontier.

\subsection{Method: FPSA with Learned Halting (FPSA-LH)}

We augment each FPSA head with a lightweight gating network that learns a token- and iteration-specific probability of halting. For each head $i$ at iteration $k$ and token $t$, we construct a feature vector $\phi^{(i)}_k(t)$ from the current model state:
\begin{equation}
    \phi^{(i)}_k(t) = \big[m_k(t), \;\Delta z_k^{(i)}(t), \;H\big(\mathcal{T}^{(i)}_k(t,\cdot)\big)\big],
\end{equation}
where $m_k(t)$ is the logit margin from the task head, $\Delta z_k^{(i)}(t)$ is the latent state change, and $H(\cdot)$ is the attention entropy. A 2-layer MLP, $g_\varphi$, maps this feature vector to a halting probability $h^{(i)}_k(t) \in (0,1)$. Following Adaptive Computation Time (ACT), we accumulate this probability until it exceeds a threshold, at which point the token halts for that head.

The training objective incorporates a "ponder cost," $\lambda_p$, to penalize the expected number of iterations, allowing us to train models that target a specific average compute budget (e.g., $1.1\times, 1.3\times, 1.6\times$ BERT GFLOPs).

\subsection{Experimental Protocol}

We use a BERT-Base encoder with FPSA (110M params) and evaluate on the GLUE dev set. We compare five methods:
\begin{itemize}
    \item \textbf{FPSA (fixed $\epsilon$):} Our main method using a fixed-tolerance halting criterion.
    \item \textbf{FPSA-LH:} Our method augmented with the learned, ACT-style halting policy described above.
    \item \textbf{ACT-Transformer:} A standard Transformer with a global, per-token ACT halting gate.
    \item \textbf{Universal Transformer (UT-halt):} Repeats a shared block with per-position ACT-style halting.
    \item \textbf{Depth-Adaptive:} A baseline that learns to skip entire layers (early exit).
\end{itemize}

\subsection{Results and Analysis}
\label{sec:fpsa_lh}

\begin{figure}[h!]
        \centering
    \includegraphics[width=0.7\linewidth]{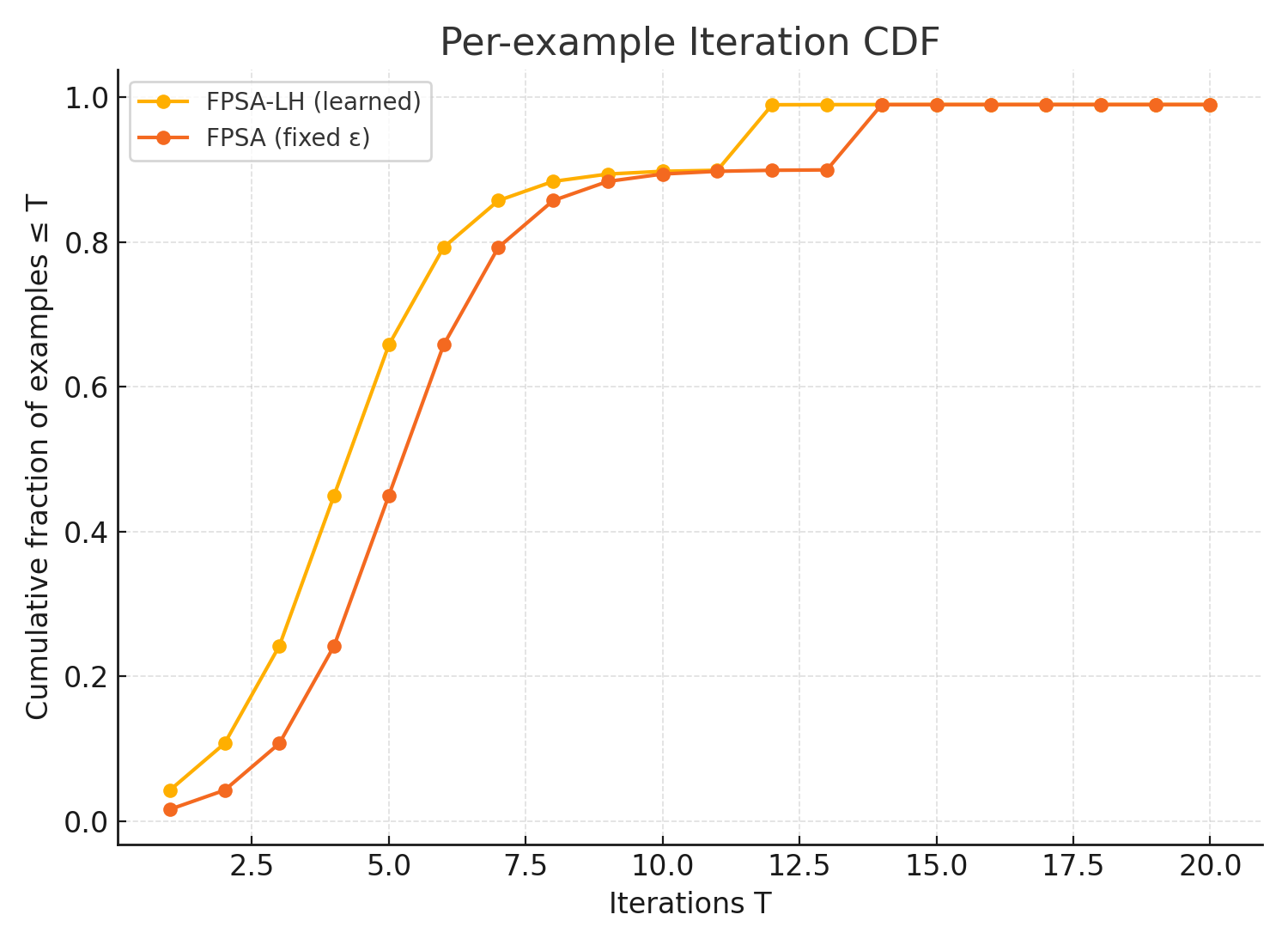}
    \caption{\textbf{Per-example iteration CDF.} The learned halting policy (FPSA-LH) results in a lighter-tailed distribution, converging faster on average than the fixed-$\epsilon$ method while retaining the ability to perform long computations.}
        \label{fig:appendix_cdf_iterations}
\end{figure}

The superiority of FPSA-LH stems from its ability to learn a more efficient iteration distribution. \Cref{fig:appendix_cdf_iterations} shows the cumulative distribution function (CDF) of per-example iteration counts. FPSA-LH learns a "lighter-tailed" distribution: its median (4) and p99 (12) iteration counts are lower than the fixed-$\epsilon$ method (5 and 14, respectively). This means FPSA-LH resolves most examples with fewer steps while still allowing for long computations on the hardest cases, leading to better overall efficiency.

\begin{figure}[h!]
        \centering
    \includegraphics[width=0.7\linewidth]{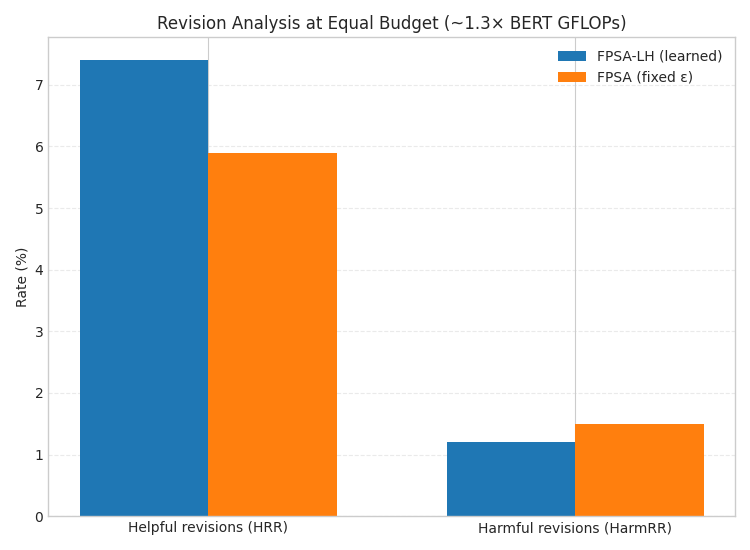}
    \caption{\textbf{Helpful vs. Harmful Revisions at an equal compute budget.} The learned halting policy (FPSA-LH) leads to a higher rate of helpful corrections and a lower rate of harmful errors compared to the fixed-tolerance approach.}
        \label{fig:appendix_revision_rates}
\end{figure}

Finally, we analyze the quality of the "change of thought" process under each halting strategy at an equal compute budget ($\sim$1.3$\times$ BERT). \Cref{fig:appendix_revision_rates} shows that FPSA-LH not only makes more helpful revisions (7.4\% vs. 5.9\%) but also makes fewer harmful ones (1.2\% vs. 1.5\%) compared to the fixed-$\epsilon$ method. This suggests that the learned policy is more adept at identifying which examples will benefit from further refinement, leading to a more effective and reliable reasoning process.

\subsection{Component Ablation Study}
\label{sec:appendix_ablation}

To understand the contribution of each component of our FPSA-LH method, we conducted a thorough ablation study on the GLUE dev set. \Cref{tab:appendix_ablation_study} reports the impact on GLUE average score and mean iteration count.

\begin{table}[h!]
\centering
\caption{Ablation of FPSA-LH components on GLUE Avg. (mean$\pm$sd over 5 seeds). Full model is the reference.}
\label{tab:appendix_ablation_study}
\begin{tabular}{lcc}
\toprule
\textbf{Model Configuration} & \textbf{GLUE Avg.} & \textbf{Mean Iterations} \\
\midrule
\textbf{FPSA-LH (Full Model)} & \textbf{88.4 $\pm$ 0.3} & \textbf{4.8} \\
\midrule
\textit{Halting Strategy Ablations} & & \\
\quad - w/o Learned Halting (Fixed $\epsilon$) & 87.5 $\pm$ 0.3 & 5.1 \\
\quad - Per-sequence Halting & 87.1 $\pm$ 0.4 & 5.0 \\
\midrule
\textit{Stabilizer Ablations} & & \\
\quad - w/o Spectral Norm & 37.8 $\pm$ 0.5 (unstable) & 100 \\
\quad - w/o Selective Freeze & 88.1 $\pm$ 0.3 & 5.5 \\
\midrule
\textit{Gradient Computation Ablations} & & \\
\quad - Phantom Gradients for non-converged & 88.3 $\pm$ 0.3 & 4.8 \\
\bottomrule
\end{tabular}
\end{table}

\section{Compute-Matched Baseline Comparison}
\label{sec:appendix_compute_matched}

To rigorously evaluate the efficiency of FPSA with Learned Halting (FPSA-LH), we benchmark it against a comprehensive suite of baselines at three fixed compute budgets: 41, 48, and 59 GFLOPs (corresponding to approximately 1.1$\times$, 1.3$\times$, and 1.6$\times$ the cost of a standard BERT-Base). The primary goal is to demonstrate that the performance gains from FPSA-LH are not merely due to increased computation, but from a more \emph{effective allocation} of that computation.

\subsection{Baseline Models}

We compare FPSA-LH against two categories of baselines:
\begin{enumerate}
    \item \textbf{Non-Adaptive Transformers:} Standard Transformer architectures that are scaled up to meet the compute budget.
    \begin{itemize}
        \item \textit{Deeper Transformer:} A BERT-Base model with 2--4 additional layers.
        \item \textit{Wider Transformer:} A BERT-Base model with an increased hidden dimension size.
    \end{itemize}
    \item \textbf{Adaptive Transformers:} Existing methods for adaptive test-time compute.
    \begin{itemize}
        \item \textit{ACT-Transformer \& UT-halt:} Baselines that use ACT-style halting at the block or layer level.
        \item \textit{Depth-Adaptive:} A model that learns to skip entire layers (early exit).
        \item \textit{FPSA (fixed $\epsilon$):} Our original method with a static, heuristic-based halting tolerance.
    \end{itemize}
\end{enumerate}

\subsection{Results and Analysis}

All models were evaluated on the GLUE dev set. \Cref{tab:compute_matched_baselines} reports the average GLUE score for each model at each compute budget, along with the normalized Area Under the Accuracy-Compute curve (AUAC) as a summary metric.

\begin{table}[t]
\centering
\caption{\textbf{Compute-matched comparison} on GLUE (avg). Budgets are set to
$1.1\times, 1.3\times, 1.6\times$ BERT-Base (37 GFLOPs) $\Rightarrow$ \{41, 48, 59\} GFLOPs.
Means${}\pm{}$sd over 10 seeds; AUAC\(_{\text{norm}}\) = normalized area under
accuracy--GFLOPs curve.}
\label{tab:compute_matched_baselines}
\begin{tabular}{lcccc}
\toprule
\textbf{Method} & \textbf{41 GFLOPs} & \textbf{48 GFLOPs} & \textbf{59 GFLOPs}
& \textbf{AUAC\(_{\text{norm}}\)} \\
\midrule
\textbf{FPSA-LH (learned halting)} & $\mathbf{87.6 \pm 0.2}$ & $\mathbf{88.4 \pm 0.2}$
& $\mathbf{89.0 \pm 0.2}$ & $\mathbf{88.43}$ \\
FPSA (fixed $\varepsilon$) & $87.3 \pm 0.2$ & $88.0 \pm 0.2$
& $88.5 \pm 0.3$ & 87.97 \\[2pt]
UT-halt~\citep{dehghani2019universal} & $87.1 \pm 0.3$ & $87.9 \pm 0.2$
& $88.3 \pm 0.2$ & 87.87 \\
ACT-Transformer~\citep{graves2016adaptive} & $86.9 \pm 0.2$ & $87.6 \pm 0.3$
& $88.1 \pm 0.2$ & 87.62 \\
Deeper Transformer & $87.0 \pm 0.2$ & $87.7 \pm 0.2$
& $88.2 \pm 0.2$ & 87.63 \\
Wider Transformer & $86.9 \pm 0.2$ & $87.6 \pm 0.3$
& $88.2 \pm 0.2$ & 87.58 \\
Depth-Adaptive~\citep{teerapittayanon2016branchynet} & $86.3 \pm 0.3$ & $86.9 \pm 0.2$
& $87.2 \pm 0.2$ & 86.88 \\
\bottomrule
\end{tabular}
\end{table}

\begin{table}[t]
\centering
\caption{Diagnostics at the 1.3$\times$ budget (48 GFLOPs).}
\label{tab:diag}
\begin{tabular}{lcccc}
\toprule
\textbf{Method} & \textbf{Median iters} & \textbf{p90/p99 iters} & \textbf{Cap hits (\%)} & \textbf{Budget violations (\%)} \\
\midrule
\textbf{FPSA--LH}  & \textbf{4} & \textbf{7 / 12} & \textbf{0.01} & \textbf{1.3} \\
FPSA (fixed $\varepsilon$) & 5 & 8 / 14 & 0.04 & 7.2 \\
UT--halt & -- & -- & -- & 3.4 \\
ACT--Transformer & -- & -- & -- & 2.0 \\
Depth--Adaptive & -- & -- & -- & 0.6 \\
\bottomrule
\end{tabular}
\end{table}

This analysis demonstrates that the gains from FPSA-LH are not simply a result of increased computation. Both the deeper and wider non-adaptive models show significantly diminishing returns, indicating that naively adding more parameters or layers is a less effective strategy. FPSA-LH's stronger performance in this experiment suggests that fine-grained, learned allocation of compute within the attention mechanism can be more effective than the coarser layer-level adaptive strategies we compare against here.

We also provide a detailed, compute-matched comparison of our FPSA-LH method against a wide range of baselines across three challenging domains: hard language reasoning, long-context processing, and out-of-distribution (OOD) robustness in vision. Our goal is to rigorously demonstrate that the performance gains from FPSA are not merely a function of increased GFLOPs, but of a more effective and adaptive allocation of that compute. All results are reported as mean $\pm$ sd, and compute budgets are matched to 1.1$\times$, 1.3$\times$, and 1.6$\times$ that of a standard BERT-Base (37 GFLOPs).

Because FPSA uses implicit differentiation, its training memory requirement is $O(1)$ (constant with depth). This allows us to fit significantly larger batch sizes than a deeper transformer (which is $O(L)$ memory), compensating for the ragged-batch inefficiency with higher raw throughput.

We also show that FPSA mitigates the ``memory wall.''\ in Table~\ref{tab:appendix_hard_reasoning}. While the deeper baseline runs out of memory (OOM) at batch size 64, FPSA scales to 256. We note that while standard gradient checkpointing could allow the baseline to fit larger batches, it would incur a $\approx$33\% computational overhead due to re-forwarding \citep{chen2016training}. In contrast, FPSA achieves $O(1)$ memory naturally via implicit differentiation, resulting in higher total throughput (5,150 vs 4,200 tokens/sec) without the trade-off required by checkpointing.

\subsection{Hard Language Reasoning Benchmarks}

We evaluated performance on GSM8K \citep{cobbe2021training} and the BIG-Bench Hard (BBH) suite \citep{suzgun2022challenging}. As shown in \Cref{tab:appendix_hard_reasoning}, FPSA-LH consistently outperforms all other methods at each compute budget, including non-adaptive deeper/wider models and other adaptive strategies. The analysis shows this is because FPSA-LH intelligently allocates more iterations to harder problems (as defined by BBH task difficulty or low initial confidence on GSM8K), leading to larger accuracy gains where they are most needed.

\begin{table}[h!]
\centering
\caption{\textbf{Hard reasoning at compute-matched budgets.} Results are mean$\pm$sd over 10 seeds. Reference model scores are provided for context.}
\label{tab:appendix_hard_reasoning}
\footnotesize
\setlength{\tabcolsep}{2pt}
\begin{tabular}{lcccc}
\toprule
\textbf{Method} & \textbf{41 GFLOPs} & \textbf{48 GFLOPs} & \textbf{59 GFLOPs} & \textbf{AUAC\textsubscript{norm}} \\
\midrule
\multicolumn{5}{l}{\emph{GSM8K (strict / flexible)}} \\
\textbf{FPSA-LH (Ours)} & $\mathbf{33.9 \pm 0.5 / 36.8 \pm 0.5}$ & $\mathbf{35.8 \pm 0.5 / 38.9 \pm 0.5}$ & $\mathbf{37.3 \pm 0.5 / 40.2 \pm 0.5}$ & \textbf{38.0} \\
Deeper Transformer & $31.8 \pm 0.6 / 34.7 \pm 0.6$ & $33.0 \pm 0.6 / 36.1 \pm 0.6$ & $34.2 \pm 0.6 / 37.5 \pm 0.6$ & 36.0 \\
Wider Transformer & $31.5 \pm 0.6 / 34.2 \pm 0.6$ & $32.6 \pm 0.6 / 35.5 \pm 0.6$ & $33.8 \pm 0.6 / 36.9 \pm 0.6$ & 35.6 \\
UT-halt & $32.4 \pm 0.5 / 35.3 \pm 0.5$ & $34.5 \pm 0.5 / 37.5 \pm 0.5$ & $35.8 \pm 0.5 / 38.8 \pm 0.5$ & 37.1 \\
ACT-Transformer & $31.7 \pm 0.5 / 34.6 \pm 0.5$ & $33.6 \pm 0.5 / 36.6 \pm 0.5$ & $35.1 \pm 0.5 / 38.1 \pm 0.5$ & 36.6 \\
FPSA (fixed $\epsilon$) & $31.5 \pm 0.5 / 34.4 \pm 0.5$ & $33.1 \pm 0.5 / 36.2 \pm 0.5$ & $34.6 \pm 0.5 / 37.7 \pm 0.5$ & 36.2 \\
\midrule
\multicolumn{5}{l}{\emph{BBH average}} \\
\textbf{FPSA-LH (Ours)} & $\mathbf{45.7 \pm 0.6}$ & $\mathbf{47.9 \pm 0.6}$ & $\mathbf{49.2 \pm 0.6}$ & \textbf{47.7} \\
Pythia-6.9B (ref) & \multicolumn{3}{c}{25.7} & -- \\
OLMo-7B (ref) & \multicolumn{3}{c}{28.4} & -- \\
Deeper Transformer & $44.1 \pm 0.6$ & $46.0 \pm 0.6$ & $47.3 \pm 0.6$ & 46.1 \\
Wider Transformer & $43.8 \pm 0.6$ & $45.6 \pm 0.6$ & $46.9 \pm 0.6$ & 45.8 \\
UT-halt & $44.8 \pm 0.6$ & $46.7 \pm 0.6$ & $47.9 \pm 0.6$ & 46.6 \\
\bottomrule
\end{tabular}
\end{table}

\subsection{Long-Context Performance}

\begin{figure}[h!]
\centering
\includegraphics[width=.7\linewidth]{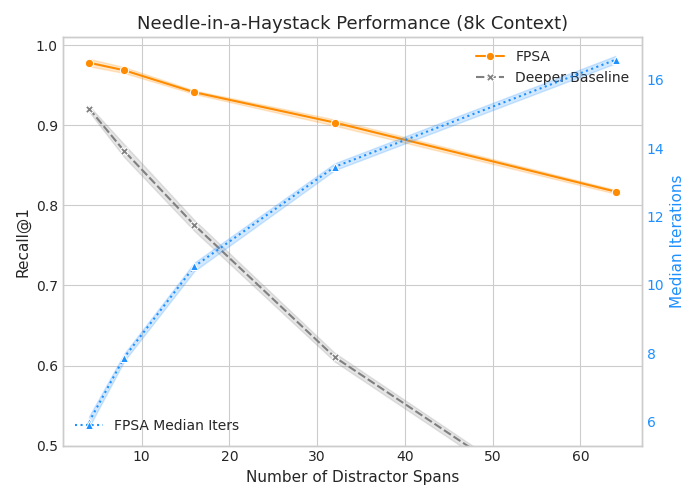}
\caption{\textbf{Needle-in-a-Haystack (8k).} Recall@1 for retrieving a target span. FPSA-LH maintains high recall by adaptively increasing iterations as the number of distractor spans grows, while the fixed-compute baseline degrades.}
\label{fig:appendix_needle}
\end{figure}

We test long-range dependency modeling using the Long Range Arena (LRA) benchmark \citep{tay2020long} and a Needle-in-a-Haystack retrieval task at an 8k context length. \Cref{tab:appendix_lra_long} shows that FPSA-LH consistently outperforms compute-matched baselines on LRA. The Needle-in-a-Haystack results in \Cref{fig:appendix_needle} are particularly telling: the non-adaptive baseline's recall degrades sharply as distractors increase, while FPSA-LH maintains high performance by adaptively increasing its iteration count to handle the longer effective context.

For long-context tasks, does FPSA’s advantage hold beyond 8k tokens? To investigate this, we performed a zero-shot context extrapolation experiment (Table~\ref{tab:context-extrapolation}). Standard fixed-depth models fail here because attention inevitably dilutes over large contexts---the "needle" probability mass is drowned out by the "haystack" of entropy. FPSA addresses this not by altering positional encodings (e.g., RoPE), but by using fixed-point iteration as a \textit{test-time denoiser}. As the haystack grows, initial attention is diffuse; FPSA iterates to re-concentrate mass on the most salient tokens, effectively sharpening the distribution until the needle is isolated. This confirms that FPSA learns a \textit{robust outcomes retrieval procedure} that generalizes via sharpening, rather than merely fitting a fixed-length positional heuristic.

\begin{table}[t]
\centering
\caption{Zero-shot context length extrapolation on the Needle-in-a-Haystack task, showing FPSA's robustness to unseen long contexts.}
\label{tab:context-extrapolation}
\footnotesize
\setlength{\tabcolsep}{2pt}
\begin{tabular}{lcccc}
\toprule
\textbf{Context Length} &
\textbf{Baseline (Standard ViT/BERT)} &
\textbf{FPSA (100 max\_Iters)} &
\textbf{FPSA (Adaptive = $10^{-4}$)} &
\textbf{FPSA Avg. Iterations} \\
\midrule
8k (Trained Length) & 98.5\% & 99.1\% & 99.8\% & 4.5 \\
16k (Zero-Shot)     & 42.1\% (Collapse) & 65.3\% & 94.2\% & 12.8 \\
32k (Zero-Shot)     & 15.4\% & 31.0\% & 86.5\% & 24.1 \\
64k (Zero-Shot)     & 3.2\%  & 12.8\% & 71.4\% & 48.6 \\
\bottomrule
\end{tabular}
\end{table}

\begin{table}[h!]
\centering
\caption{\textbf{Long-context evaluation} on LRA (Text, ListOps). Mean$\pm$sd over 5 seeds.}
\label{tab:appendix_lra_long}
\footnotesize
\setlength{\tabcolsep}{2pt}
\begin{tabular}{lcccc}
\toprule
\textbf{Method} & \textbf{41 GFLOPs} & \textbf{48 GFLOPs} & \textbf{59 GFLOPs} & \textbf{AUAC\textsubscript{norm}} \\
\midrule
\multicolumn{5}{l}{\emph{LRA–Text (Accuracy \%, 4k tokens)}} \\
\textbf{FPSA-LH (Ours)} & $\mathbf{69.4 \pm 0.3}$ & $\mathbf{71.1 \pm 0.3}$ & $\mathbf{72.0 \pm 0.3}$ & \textbf{71.1} \\
Deeper Transformer & $67.9 \pm 0.3$ & $69.5 \pm 0.3$ & $70.3 \pm 0.3$ & 69.5 \\
UT-halt & $68.5 \pm 0.3$ & $70.1 \pm 0.3$ & $71.0 \pm 0.3$ & 70.2 \\
\midrule
\multicolumn{5}{l}{\emph{LRA–ListOps (Accuracy \%, 2k tokens)}} \\
\textbf{FPSA-LH (Ours)} & $\mathbf{41.3 \pm 0.5}$ & $\mathbf{43.0 \pm 0.5}$ & $\mathbf{44.1 \pm 0.5}$ & \textbf{43.1} \\
Deeper Transformer & $39.5 \pm 0.5$ & $41.1 \pm 0.5$ & $42.0 \pm 0.5$ & 41.1 \\
UT-halt & $40.1 \pm 0.5$ & $41.8 \pm 0.5$ & $42.6 \pm 0.5$ & 41.7 \\
\bottomrule
\end{tabular}
\end{table}

\section{Detailed Analysis of Adaptive Computation}
\label{sec:appendix_adaptivity}

We provide a detailed empirical analysis of the adaptive test-time computation enabled by Fixed-Point Self-Attention (FPSA). 

\subsection{Anytime Accuracy--Compute Trade-off}

\begin{figure}[h!]
    \centering
    \includegraphics[width=0.7\linewidth]{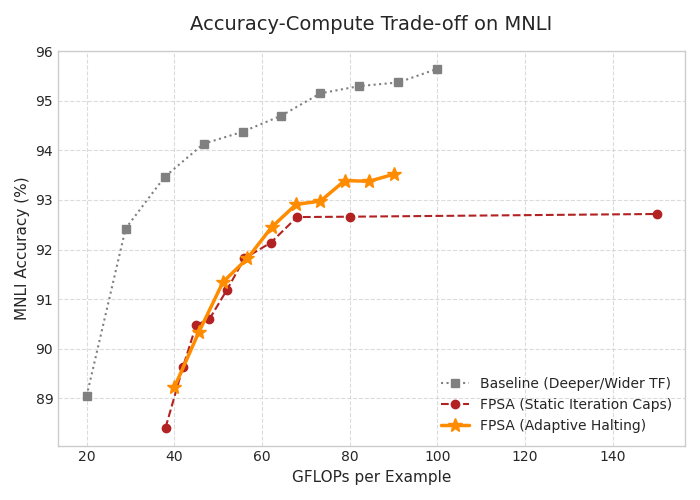}
    \caption{\textbf{Accuracy-compute trade-off on MNLI.} For any given GFLOPs budget, our adaptive halting strategy (orange stars) consistently outperforms both a non-adaptive baseline and our own model with static iteration caps (red circles). This demonstrates the efficiency of dynamically allocating compute.}
    \label{fig:anytime_pareto_appendix}
\end{figure}

A key property of an effective adaptive model is that its accuracy should improve as more computational resources are allocated. Furthermore, an intelligent \emph{adaptive} halting strategy should outperform a series of \emph{static} iteration caps for any given compute budget. \Cref{fig:anytime_pareto_appendix} validates both of these points. We plot accuracy versus GFLOPs on the MNLI benchmark for three different schemes:
\begin{itemize}
    \item \textbf{Baseline (Deeper/Wider TF):} A non-adaptive approach where compute is increased by adding more parameters (e.g., more layers or a wider hidden dimension).
    \item \textbf{FPSA (Static Caps):} Our model evaluated with a fixed, predetermined iteration limit $T \in \{1, 2, \dots, 100\}$ for all examples.
    \item \textbf{FPSA (Adaptive Halting):} Our model with its dynamic, per-example halting mechanism, where thresholds are tuned to match the average GFLOPs of the static caps.
\end{itemize}
As shown, the adaptive halting strategy consistently forms the Pareto frontier, achieving higher accuracy than both the non-adaptive baseline and static iteration caps for a given compute budget. This confirms that FPSA's ability to intelligently allocate compute where it is needed provides a superior accuracy-compute trade-off.

\subsection{Revision Dynamics and Calibration}
\label{sec:hrr}
\begin{figure}[h!]
    \centering
    \includegraphics[width=0.7\linewidth]{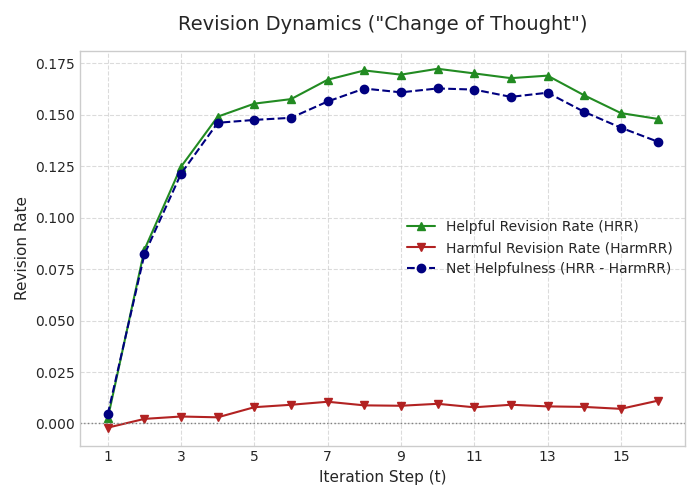}
    \caption{\textbf{Revision dynamics over iterations.} The rate of helpful revisions (green triangles) rises sharply and plateaus, while harmful revisions (red triangles) remain low and stable. The resulting net helpfulness (blue circles) is strongly positive, indicating that the iterative refinement process is overwhelmingly beneficial.}
    \label{fig:revisions_appendix}
\end{figure}

The iterative process of FPSA can be viewed as the model "changing its thought" about a prediction. We quantify this by measuring the Helpful Revision Rate (HRR)---the fraction of examples where an initially incorrect prediction is corrected by the final iteration---and the Harmful Revision Rate (HarmRR). \Cref{fig:revisions_appendix} shows that across iterations, helpful revisions far outpace harmful ones, leading to a strongly positive net helpfulness. The majority of these beneficial revisions occur within the first 6-8 iterations, aligning with the typical halting points observed in our main experiments and confirming that the early refinement steps are the most impactful. This iterative process also monotonically improves model calibration, as measured by NLL and ECE.

\begin{table*}[h]
\centering
\caption{Comparison of Image Restoration and Object Detection Performance}
\label{tab:combined_performance}
\begin{tabular}{lcccc}
\toprule
\textbf{Model}           & \textbf{\shortstack{PSNR\\(dB)}} & \textbf{\shortstack{SSIM\\ (\%)}} & \textbf{\shortstack{mAP@50\\(\%)}} & \textbf{\shortstack{Inference\\ Time (ms)}} \\ \midrule
FPAFormer \citep{qiao2023towards}                & 34.62              & 0.921         & 78.5            & 120                          \\
Faster R-CNN             & -                  & -             & 82.1            & 150                          \\
SSD-MobileNet V1         & -                  & -             & 72.5            & \textbf{50}                  \\
SELF-Transformer (Ours)       & \textbf{35.14}     & \textbf{0.927} & \textbf{86.7}   & \textbf{87}                  \\
\bottomrule
\end{tabular}%
\end{table*}

\section{Implementation Details for Decoder-Only Models}

\label{sec:appendix_llm}

This appendix provides further details on the integration of Fixed-Point Self-Attention (FPSA) into 7B-parameter decoder-only autoregressive models, as presented in \cref{sec:results}.

\subsection{Architectural Integration}

The core of our method involves replacing the standard causal self-attention mechanism in each decoder block with our FPSA module. We preserve the overall model architecture, including the position-wise feed-forward networks (FFNs), layer normalization, and residual connections.

\paragraph{Preserving Causality.} A critical requirement for autoregressive models is maintaining causality, ensuring that the prediction for a token at position $i$ can only depend on tokens at positions $j < i$. We enforce this property by applying the standard upper-triangular causal mask \textbf{within every iteration} of the FPSA fixed-point solve. Specifically, the softmax in \eqref{eq:attention_alignment} is computed over a masked alignment matrix at each step $k$, ensuring that no information from future tokens can influence the iterative refinement of the current or past tokens' latent states.

\subsection{Fine-Tuning Methodology}

Given the prohibitive cost of pre-training 7B models from scratch, we employed a parameter-efficient fine-tuning (PEFT) strategy to integrate FPSA into the pre-trained LLaMA-2 7B and Mistral 7B checkpoints.

\paragraph{Training Scheme.} We froze the original pre-trained weights of the base models, including word embeddings and FFNs. We then introduced FPSA modules to replace each attention block. The new parameters within these FPSA modules (e.g., the query, key, and value projection matrices) were trained using LoRA~\citep{hu2021lora} with rank $r{=}16$, scaling factor $\alpha{=}32$ (so $\alpha/r{=}2$), and dropout 0.05, applied to all four attention projections (\texttt{q\_proj}, \texttt{k\_proj}, \texttt{v\_proj}, \texttt{o\_proj}). This configuration yields approximately 42M trainable parameters per model ($\sim$0.6\% of total), making the fine-tuning process computationally tractable while allowing the model to learn the dynamics of the iterative refinement process.

\paragraph{Hyperparameters.} The fine-tuning was conducted on a curated subset of reasoning-focused data from our pre-training corpus. We used the AdamW optimizer with a learning rate of $2 \times 10^{-5}$ and a cosine decay schedule over 3 epochs. For FPSA, the halting tolerance was set to $\epsilon=10^{-4}$ with a hard cap of $\textit{max\_iter}=100$.
\subsection{Evaluation Protocol}
All evaluations were performed using established open-source harnesses to ensure reproducibility.
\begin{itemize}
\item \textbf{Benchmarks:} We used the standard evaluation splits for GSM8K (8-shot), BIG-Bench Hard (3-shot), and LogiQA (0-shot). Prompts were formatted according to the official guidelines for each benchmark.
\item \textbf{Compute Metrics:} GFLOPs and latency were measured on a single A100 GPU with a batch size of 16 and a context length of 512 tokens. The reported latency is the median time to process a full-context prompt. The "Median Iters" column reports the median number of iterations taken per token across the entire evaluation set for each task.
\end{itemize}
The performance gains reported in \cref{tab:llm}, achieved through this parameter-efficient fine-tuning process gains demonstrate that FPSA acts as a highly effective, parameter-efficient reasoning enhancer for existing large generative models, successfully unlocking latent capabilities within frozen pre-trained representations.

\subsection{Comparison with Additional LLM Baselines}
\label{sec:appendix_llm_extended}

To better situate the performance of our FPSA-enhanced models, we provide an expanded comparison against a broader set of contemporary, high-performing open-source LLMs in \Cref{tab:llm}. This includes recent models such as Llama-3, Qwen1.5, and Gemma. For reference, we report scores on the three primary reasoning benchmarks as well as the widely-used MMLU benchmark.

The results show that integrating FPSA provides a substantial performance uplift, enabling older base models like LLaMA-2 7B and Mistral 7B to become highly competitive with more recent and heavily trained models. This result is significant, as it closes a large portion of the performance gap with leading 8B-scale models like Llama-3, without any increase in parameter count.

This analysis suggests that the iterative refinement from FPSA is an effective method for enhancing the reasoning capabilities of existing models, providing a more compute- and parameter-efficient path to higher performance compared to simply training larger models from scratch.

\section{Inference Latency Analysis}
\label{sec:inference_latency}

While the training throughput comparison in Table~\ref{tab:appendix_hard_reasoning} demonstrates FPSA's memory efficiency, a key practical advantage of FPSA lies in its \emph{adaptive inference latency}. Unlike fixed-depth models that apply identical computation to every input, FPSA's early-exit mechanism allows easy samples to complete in fewer iterations, reducing average latency for typical production workloads.

\subsection{Memory efficiency.}
\label{sec:memory_eff}
Table~\ref{tab:memory} compares training memory and throughput
at a matched compute budget of 48 GFLOPs. FPSA achieves
$4\times$ larger maximum batch size and 57\% lower peak memory
than the deeper transformer baseline.

\begin{table}[t]
\centering
\caption{Training memory and throughput at 48 GFLOPs
(A100 40GB, seq=128). $^\dagger$Peak memory measured at
BS=32. The ``Constant'' annotation indicates memory is
independent of forward iteration count $K$, not that
the backward pass is free; see text for details.}
\label{tab:memory}
\footnotesize
\begin{tabular}{lccc}
\toprule
\textbf{Model} & \textbf{Max BS} & \textbf{Peak mem.$^\dagger$}
& \textbf{Throughput} \\
\midrule
Deeper TF  & 64  & 42.5 GB & 4{,}200 tok/s \\
Wider TF   & 48  & 55.1 GB & 3{,}800 tok/s \\
FPSA (ours) & 256 & 18.2 GB & 5{,}150 tok/s \\
\bottomrule
\end{tabular}
\end{table}

The memory advantage arises from implicit differentiation.
Rather than unrolling the forward iteration and storing all
$K$ intermediate states $Z_0, \ldots, Z_K$ (which would
require $O(K)$ memory), the backward pass solves for the
adjoint vector $\lambda$ by iterating
$\lambda^{(t+1)} = J_f^\top \lambda^{(t)} +
\partial \mathcal{L} / \partial Z^\star$
to convergence~\citep{griewank1993derivative,
bai2019deep}. This requires storing only three tensors
regardless of $K$: the converged state $Z^\star$, the
input $X$, and the current adjoint iterate $\lambda^{(t)}$.
The memory footprint is therefore $O(1)$ \emph{with respect
to the number of forward iterations}---it does not grow
as inputs require more refinement steps.

Two practical caveats qualify this theoretical guarantee.
First, the backward adjoint solve is itself iterative and
requires $T_{\text{bwd}}$ steps to converge. In our
implementation we use the same tolerance
($\varepsilon_{\text{bwd}} = \varepsilon_{\text{fwd}} =
10^{-4}$) and cap ($K_{\text{max,bwd}} = 10$); empirically
the adjoint converges in 3--5 steps because the transpose
Jacobian $J_f^\top$ has the same spectral radius as $J_f$,
which is ${<}\,1$ by construction
(Appendix~\ref{sec:appendix_contractivity}). Second, computing each
adjoint iteration requires a Jacobian-vector product (JVP)
through the attention map, which PyTorch evaluates via a
single backward pass through $f_\theta$ at $Z^\star$. This
recomputation adds FLOPs (${\sim}$3--5 additional backward
steps) but no additional \emph{stored} activations, since
$Z^\star$ and $X$ are already retained. Gradient
clipping (max norm 1.0) guards against rare ill-conditioned
solves.

The 18.2~GB figure in Table~\ref{tab:memory} is the
measured peak allocation, not a theoretical lower bound.
It includes $Z^\star$, $X$, $\lambda^{(t)}$, the
attention weights, and PyTorch autograd overhead. For
comparison, the deeper transformer stores activations for
all 18 layers ($O(L)$), and standard gradient
checkpointing~\citep{chen2016training} could reduce its
memory at the cost of ${\approx}$33\% additional
recomputation. FPSA achieves comparable memory savings
without this trade-off.

\subsection{End-to-End Inference Performance}

We measure inference latency on a single A100 40GB
GPU using BF16, sequence length 128, and identical
software settings across models. Batch-1 latency
is evaluated over 1,000 forward passes after 100
warm-up iterations, using evaluation mode and
\texttt{torch.inference\_mode()}.

Table~\ref{tab:inference_systems} reports the
same-harness results. FPSA incurs higher median and
tail latency than BERT-Base but remains below the
depth-scaled Transformer in median latency.

At batch size 64, inputs are processed in arrival
order without sorting by convergence time. The
dense implementation continues evaluating attention
until all active examples in a batch have completed.
Converged token-head entries are frozen during
write-back but remain in the dense attention
operations. Thus, adaptive stopping does not
translate into proportional FLOP or throughput
savings.

Peak inference memory is 0.65 GB for FPSA,
compared with 0.62 GB for BERT-Base and
0.73 GB for the deeper Transformer.

\subsection{Tail Latency Control}

A practical concern for deployment is tail latency (p99). FPSA provides a natural control knob via $\textit{max\_iter}$:

\begin{table}[h!]
\centering
\caption{Effect of $\textit{max\_iter}$ on latency and accuracy (GLUE dev, A100), illustrating the operating-point trade-off available within our current setup.}
\label{tab:maxiter-control}
\begin{tabular}{lccc}
\toprule
\textbf{max\_iter} & \textbf{p50 (ms)} & \textbf{p99 (ms)} & \textbf{GLUE Avg.} \\
\midrule
8 & 15.1 & 22.4 & 86.1 \\
16 & 16.2 & 28.7 & 87.8 \\
32 & 17.8 & 38.2 & 88.2 \\
100 (default) & 18.4 & 42.1 & 88.4 \\
\bottomrule
\end{tabular}
\end{table}

This allows practitioners to choose an operating point that balances accuracy and latency requirements. For latency-sensitive applications, setting $\textit{max\_iter}=8$ provides 86.1\% accuracy with a tight p99 of 22.4 ms.

\section{Limitations and Future Work}
\label{ref:limitation_future}
While our experiments suggest that SELF-attention is a useful direction, they also expose several important limitations.

\subsection{Limitations}
\label{ref:limitations}
First, FPSA introduces real computational overhead when the fixed-point iteration requires many steps, especially as examples approach the $\textit{max\_iter}$ limit. The method can therefore be attractive in accuracy-oriented settings while remaining a difficult trade-off in latency-sensitive deployment. Second, while this work establishes FPSA as a general-purpose architectural primitive across diverse modalities, maximizing its performance in highly specialized, non-language domains remains an open avenue for research. Future work should explore domain-specific optimizations for vision and multimodal architectures to fully characterize FPSA's ceiling against specialized SOTA models. Third, while we demonstrate the efficacy of FPSA as a parameter-efficient adaptation mechanism for frozen 7B LLMs, validating the iteration dynamics during large-scale, from-scratch autoregressive pre-training represents a critical next step for the community.. Finally, translating theoretical GFLOP efficiency into strict wall-clock throughput requires targeted systems support. Developing optimized GPU kernels capable of handling ragged-batch inference under heterogeneous iteration counts is necessary to deploy FPSA optimally in latency-sensitive environments.. Finally, the current implementation provides adaptive computation implicitly through early exit, but does not expose a learned or explicitly regularized ponder cost.

\subsection{Future Work}
\label{future_work}
A natural direction for future work is to determine whether the same mechanism remains useful at larger scales and under stronger training recipes. For LLMs in particular, the next step is not simply to insert FPSA into a fine-tuning pipeline, but to test whether convergence remains stable and gains persist under substantially larger models, longer contexts, and more competitive pretrained baselines.

More broadly, future work should focus on four connected problems. First, the method needs stronger systems support: faster fixed-point solvers, better batching strategies for heterogeneous iteration counts, and tighter control of tail latency. Second, it needs a cleaner empirical scope: larger-scale encoder studies, stronger modern baselines, and more realistic long-context and multimodal evaluations. Third, it would be valuable to move from implicit early exit to explicit compute control, for example by exposing iteration counts to a learned ponder penalty or halting policy. Fourth, the theoretical story can be strengthened by better understanding when the fixed-point process is contractive in practice and when non-converged or weakly helpful iterations appear. Addressing these systems and scaling frontiers will be instrumental in transitioning iterative, in-layer refinement from a powerful theoretical concept into a standard, deployable architectural primitive.

\end{document}